\documentclass[lettersize,journal]{IEEEtran}
\usepackage{amsmath,amsfonts}
\usepackage{algorithmic}
\usepackage{algorithm}
\usepackage{array}
\usepackage{textcomp}
\usepackage{stfloats}
\usepackage{url}
\usepackage{verbatim}
\usepackage{graphicx}
\usepackage{cite}
\usepackage{mathtools}
\usepackage{tensor}
\usepackage{multirow}
\usepackage{booktabs}
\usepackage{gensymb}
\usepackage{amssymb}
\usepackage{subfigure}
\usepackage{siunitx}
\usepackage[colorlinks,linkcolor=black]{hyperref}
\usepackage[switch]{lineno}
\begin{document}

\title{
Loop Closure Detection Based on Object-level Spatial Layout and Semantic Consistency}

\author{\IEEEauthorblockN{Xingwu Ji, Peilin Liu*, Haochen Niu, Xiang Chen, Rendong Ying, Fei Wen
}


\thanks{The authors are with the Brain-Inspired Application Technology Center (BATC), School of Electronic Information and Electrical Engineering, Shanghai Jiao Tong University, Shanghai 200240, China. (email: \{jixingwu, liupeilin, haochen\_niu, ChenXiang\_hit, rdying, wenfei\}@sjtu.edu.cn )}
}



\maketitle

\begin{abstract}
Visual simultaneous localization and mapping (SLAM) systems face challenges in detecting loop closure under the circumstance of large viewpoint changes. In this paper, we present an object-based loop closure detection method based on the spatial layout and semantic consistency of the 3D scene graph. Firstly, we propose an object-level data association approach based on the semantic information from semantic labels, intersection over union (IoU), object color, and object embedding. Subsequently, multi-view bundle adjustment with the associated objects is utilized to jointly optimize the poses of objects and cameras. We represent the refined objects as a 3D spatial graph with semantics and topology. Then, we propose a graph matching approach to find correspondence objects based on the spatial layout and semantic property similarity of vertices' neighbors. Finally, we jointly optimize camera trajectories and object poses in an object-level pose graph optimization, which results in a globally consistent map. Experimental results demonstrate that our proposed data association approach can construct more accurate 3D semantic maps, and our loop closure method is more robust than point-based and object-based methods in circumstances with large viewpoint changes.
\end{abstract}


\section{Introduction}
\IEEEPARstart{V}{isual} simultaneous localization and mapping (SLAM) is a widely used technique in the field of robotics for accurate pose localization and drift-free map construction. In SLAM systems, loop closure detection is a fundamental component that has been commonly used to reduce drift errors.  Typically, the classical visual SLAM systems such as 
FAB-MAP \cite{fab-map}, VINS-Mono \cite{qin2017vins}, and ORB-SLAM3 \cite{ORBSLAM3_TRO} use loop closure engines that rely on local appearance-based features \cite{VALGREN2010149, orb, sift, surf_2006} and Bag-of-Words \cite{BOW, DBOW2} extracted from the features to recognize revisited places and reduce drift errors. Though efficient, the methods based on local appearance features face challenges in many practical applications due to imaging system noise and environment appearance variation, especially when detecting loop closure under the circumstance of large viewpoint changes.

\begin{figure}[ht]
    \centering
    \includegraphics[width=0.4\textwidth]{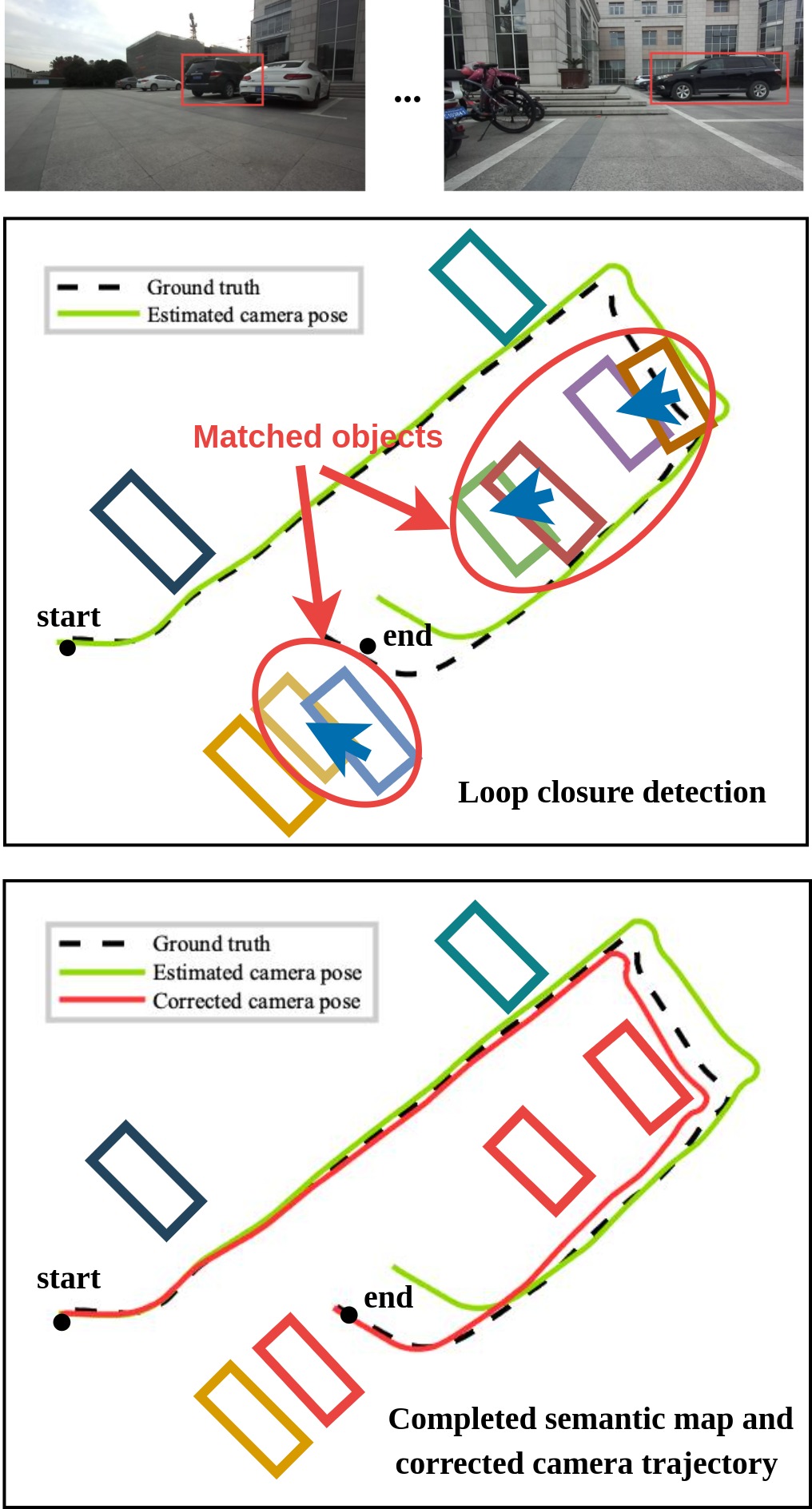}
    \caption{Object-level loop closure detection based on semantic mapping. Top: a loop image and a current image from the HALV dataset. Middle: semantic map from bird's eye view and loop closure detection. Bottom: completed semantic map and corrected camera trajectory (red line) based on the alignment of correspondence objects (red rectangles).}
    \label{fig:example}
\end{figure}

Typically, in classical visual SLAM systems, loop closure detection relies on local appearance-based features obtained through clustering and encoding image block intensities, based on which to represent the environment as a database of images. In the localization and mapping procedure, an input query image is matched with this database to retrieve the most similar candidates to detect whether it is a revisited scene. Loop closure detection methods based on these local appearance features are generally efficient and perform well in the cases of small scene appearance variation and small viewpoint change. However, in the cases of large scene appearance variation and large viewpoint change, such methods deteriorate dramatically. Specifically, when the camera’s view angle changes in the scene, the classical methods tend to fail since the same local features are no longer visible \cite{li_semantic_2019}.

In contrast to local appearance-based features, geometry and semantic features are more invariant with respect to the viewpoints and appearance changes in the environment. Recently, many methods have been proposed to incorporate high-level features into SLAM systems to improve pose estimation accuracy and robustness, e.g., OrcVIO \cite{orcvio}, CubeSLAM\cite{cubeslam}, EAO-SLAM\cite{eao-slam}, and SO-SLAM \cite{so-slam}. It has been shown that exploiting high-level features can significantly improve the robustness and accuracy of data association, especially in the condition of partial observations and occlusions. However, the spatial relationship and topological structure information of the environment, which is beneficial to achieving consistent loop closure detection and global localization, has been ignored in these methods.

In the paper, we propose a novel loop closure detection approach based on the consistency of spatial and geometric structure. Firstly, we employ the CNN-based detector \cite{liu_ground-aware_2021} to extract the objects in the environment as high-level semantic landmarks and propose an object-level data association method to construct the object-level semantic map. We further utilize the bundle adjustment (BA) algorithm \cite{ba} to jointly optimize the poses of objects and cameras in the multi-view of the objects. Then, we match the spatial layout and semantic properties to detect loop closure based on the object’s surrounding topological structures in the local and global maps. Experimental results have shown that our method can recognize a loop closure even when the viewpoint changes over 125 degrees. Finally, we present an object-level pose optimization to maintain the consistency of the semantic map and the camera trajectory. An illustration of our method is given in Fig. \ref{fig:example}.

The contributions are summarized as follows: 
\begin{itemize} 
    \item A loop closure detection method based on topological structure and geometry matching in the 3D scene graph, which leverages the information of objects and their neighbors to construct 3D topology structures with spatial layout consistency and semantic properties consistency.

    \item An object-level data association method, which incorporates semantic label, bounding box IoU, object color encoding, and object-level embedding into the similarity calculation. The method results in the construction of a 3D semantic map that is robust to camera viewpoint changes.
    
    \item A complete visual SLAM system, which is verified under benchmark datasets with various viewpoint conditions. Experimental results show that, in the condition of large viewpoint changes, our system is superior over existing appearance-based and object-based SLAM systems in terms of accuracy and robustness.
    
\end{itemize}

The rest of the paper is organized as follows. In Section \ref{sec:Related Work}, we review the relevant literature. Section \ref{sec:object-level_semantic_mapping} explains the object-level semantic mapping in detail. Section \ref{sec:loop_closure_with_semantic_landmarks} introduces the loop closure detection and object-level pose graph optimization. In Section \ref{sec:results}, we evaluate our system in comparison with other state-of-the-art algorithms on public datasets and our high-accuracy datasets with large viewpoint changes (HALV). Finally, a summary of this work is concluded in Section \ref{sec:conclusion}.

\section{Related Work}\label{sec:Related Work}
\subsection{Object SLAM}
Recently, many visual SLAM systems have been developed, which rely on local appearance-based features and image intensities \cite{gong_graph_2021, miao_univio_2022}. These systems can achieve accurate mapping and localization in general environments. However, they are still vulnerable to significant changes in illumination and view angle when used in outdoor environments. Compared to the low-level features, object features are more stable and can provide richer semantic information. With the rapid advancements in machine learning, there is a growing interest in replacing low-level features with objects based on CNN models \cite{liu_ground-aware_2021, rcnn, MaskRCNN} in visual SLAM systems.

For instance, Salas et al. \cite{salas-moreno_slam_2013} introduced a SLAM framework called SLAM++, which uses dense object models to assist in mapping. However, the method relies on prior geometric information and an object database. In contrast, Grinvald et al. \cite{grinvald_volumetric_2019} and Mccorman et al. \cite{mccormac_fusion_2018} utilize voxel-based representation to model surrounding objects without prior information. However, the requirement for dense voxel models limits the applicability of the methods. Alternatively, some methods use general three-dimensional geometric structures, such as spheres \cite{frost_recovering_2018, bowman_probabilistic_2017} and cuboids \cite{cubeslam} to represent objects in the environment. 
For example, Frost et al. \cite{frost_recovering_2018} incorporated object centroids as point clouds into the camera pose estimation process, while Bowman et al. \cite{bowman_probabilistic_2017} included additional object positions in the SLAM framework pipeline. However, they did not explore and look at more signals, i.e., scales and poses.
CubeSLAM \cite{cubeslam} is a real-time monocular SLAM system that utilizes 3D cuboids with scales and poses to construct accurate environmental models, incorporating cuboid objects within the scene. However, the data association of the method still relies on feature point matching, which tracks feature points to match the bounding box of an object and may also lead to difficulties when numerous similar objects are present in the scene.

In this work, we introduce a novel object-level data association method that extracts object features from the environment using a CNN-based model \cite{liu_ground-aware_2021} and tracks them by incorporating the semantic labels, bounding box IoU, object color encoding, and object-level embedding. To achieve accurate object localization, we jointly optimize and refine the poses of cameras and objects to minimize geometric and location error. In the experiments, We show a semantic map with precise 3D objects and a globally consistent camera trajectory. 

\begin{figure*}[ht]
    \centering
    \includegraphics[width=\textwidth]{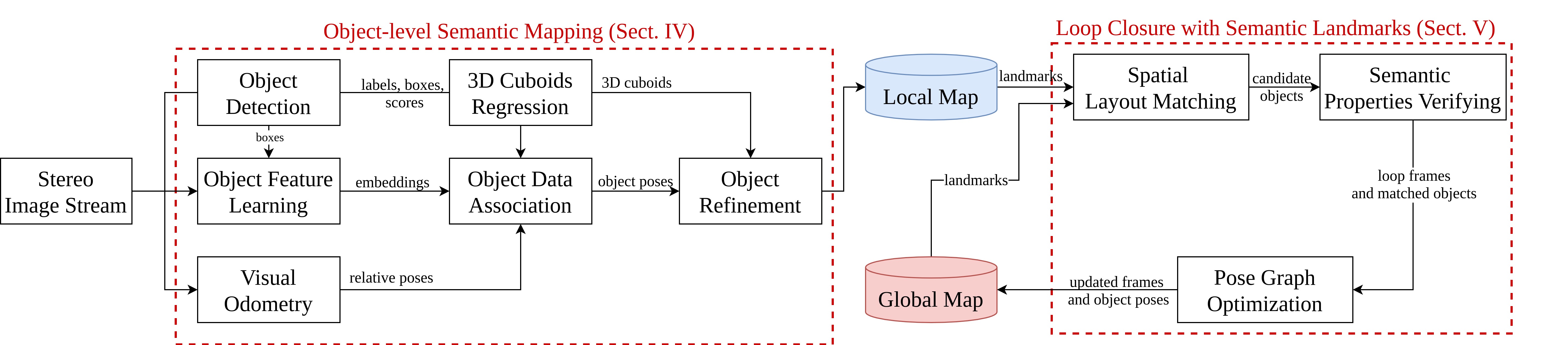}
    \caption{Overall architecture of the proposed system with object-level semantic mapping and semantic landmarks based on loop closure.}
    \label{fig:overview}
\end{figure*}

\subsection{Loop Detection}
Loop closure is an essential component of simultaneous localization and mapping, which depends on the recurrent observation of landmarks to mitigate the cumulative errors that arise from the robot's motion. Conventional methods \cite{ORBSLAM3_TRO, qin2019fusion} typically extract and encode local point features to establish a database of images, and then retrieve images in the database. However, these point-based methods are sensitive to changes in viewpoint and environmental appearance.

Recently, many works seek to incorporate high-level semantic features to address the loop closure problem in cases with significant changes in appearance. These works represent the environment as a graph structure, and reformulate the loop closure detection as a graph matching problem. Moreover, some methods represent each image with identified objects as a graph and encode each object within the graph. For example, Kuan et al. \cite{aircode} gather and encode the feature points in the region of the objects. Kim et al. \cite{kim_closing_2022} integrate visual points into semantic objects as kernel nodes and introduce a graph kernel algorithm to match query and target graphs. Nevertheless, when the viewpoint changes significantly and the objects are occluded, the same object is represented inconsistently in both graphs, leading to a decline in the performance of these methods.

On the other hand, some methods leverage oriented objects in time-continuous images to construct global graphs. For instance, X-view \cite{gawel_x-view_2018} creates 2D semantic graphs using image sequences with object segmentation. In the graphs, vertices represent semantic blobs, and edges indicate proximity relations. Graph matching is achieved by comparing random walk descriptors between vertices. These descriptors contain topological information of the semantic graph, making them resistant to seasonal and significant viewpoint changes. Moreover, some methods extend this work to edit distance matching \cite{lin_topology_2021} and semantic histogram-based matching \cite{guo_semantic_2021}. Furthermore, the methods \cite{qian_towards_2022, semanticloop} utilize 3D graph co-visibility to match objects in query and global graphs. However, when there are multiple objects of the same category in the graph or the graphs have similar topological structures, random walk-based and 3D co-visibility-based methods may both incorrectly identify objects and decrease the performances.

In this work, we present a novel loop closure detection method based on semantic topology graph matching. The method matches correspondence objects with the spatial layout of the graphs, followed by semantic properties verifying to ensure accurate matching. Our system is a complete pipeline integrated into a classical visual SLAM framework, as depicted in Fig. \ref{fig:overview}. In the experiments, we demonstrate that our method is more accurate and robust in achieving globally consistent localization when the camera's view angle changes over 125 degrees.

\section{Object-level Semantic Mapping} \label{sec:object-level_semantic_mapping}
Using stereo RGB streams as input, we utilize off-the-shelf stereo odometry to acquire primitive relative poses. Concurrently, a separate thread processes the RGB frame using an object detection network to identify semantic labels, and extract 2D bounding boxes and 3D cuboid proposals. These bounding boxes are then fed into an instance-feature learning network to extract the object-level embeddings. Additionally, objects that are either out of size or far from the camera are filtered out to improve the robustness of subsequent processing. Leveraging the object semantic information, the object data association algorithm performs to match the detected objects in the local map with the objects in the global map. In case of no object in the global map is matched with the detected ones, we create a new cuboid and append it to the global map. When an object is associated, we apply the object pose refinement algorithm to fuse the new measurements into the cuboid and employ local bundle adjustment approach \cite{ba} to jointly optimize the poses of the objects and cameras.

\subsection{Object Semantic Information Extraction} \label{sec:generation} 


\textit{1) Geometry information:}
The geometric information includes semantic labels, object sizes, 2D bounding boxes, and 3D cuboid proposals. The process of regressing 2D bounding boxes and 3D cuboids from an image involves the use of the 3D object detector, such as VisualDet3D \cite{liu_ground-aware_2021}. The detector uses ResNet-101 \cite{he_deep_2016} as the backbone network to extract high-level features and only takes features at scale of $1/16$. The feature map is then fed into the classification branch and regression branch to predict the parameters of the bounding boxes and cuboids. The object detector \cite{liu_ground-aware_2021} was trained using the KITTI object dataset \cite{inproceedingskitti} for the purpose of object extraction.

\begin{figure}[tp]
    \centering
    \includegraphics[width=0.45\textwidth]{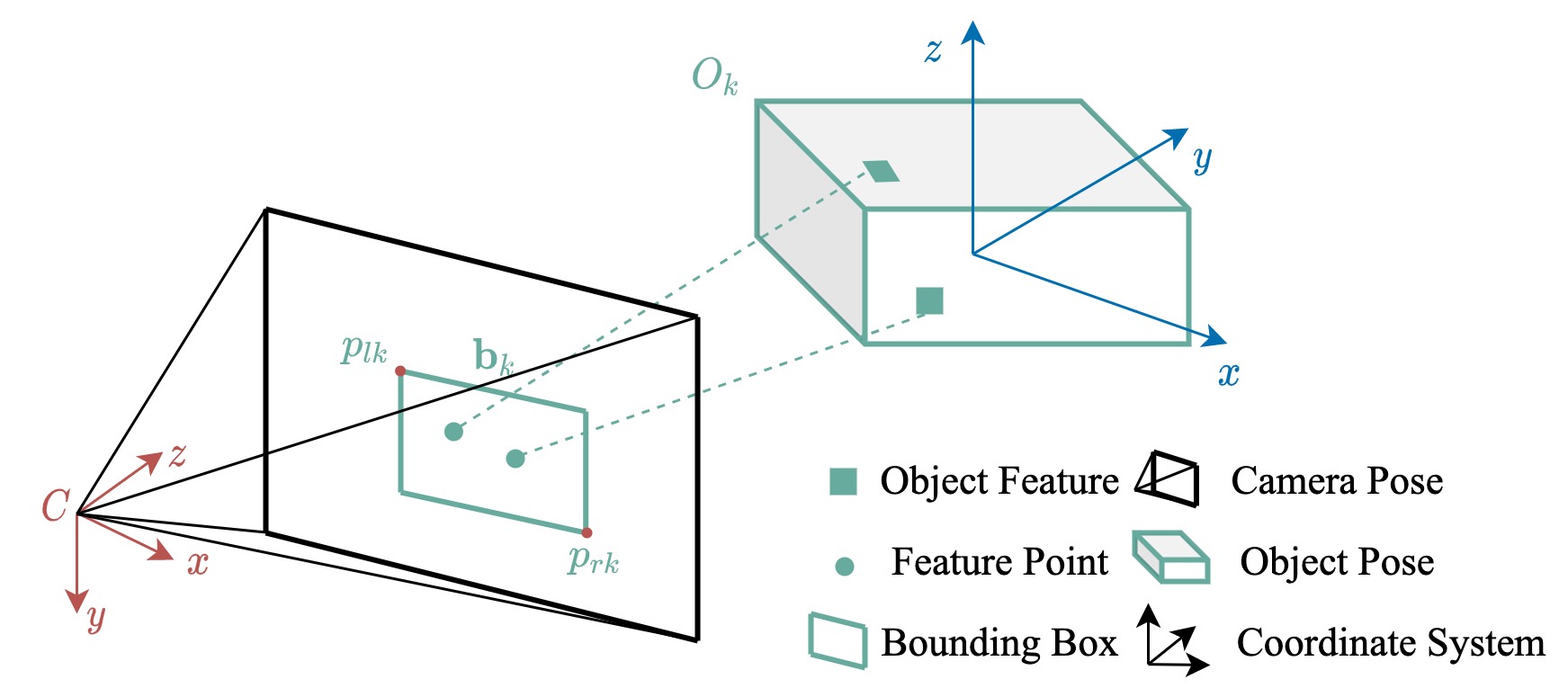}
    \caption{The visualization of the symbols used in our system}
    \label{fig:coordinate_system}
\end{figure}

We follow the idea from YOLOv3 \cite{redmon_yolov3_2018} to densely define the anchors. Each anchor on the image acts as a proposal of an object in 3D. A 3D anchor consists of a 2D bounding box parameterized by $\mathbf{b}_k=[p_{lk}, p_{rk}]$, where $p_{lk}$ and $p_{rk}$ are the left-top and right-bottom coordinates. The 3D cuboids are represented by 9 degree-of-freedom (DoF) parameters: 3 DoF position vector $\mathbf{t}=[t_x, t_y, t_z]$, 3 DoF rotation matrix $R$ and 3 DoF dimension vector $\mathbf{d}=[d_x, d_y, d_z]$. The object poses are denoted as $T_{co}$ in the camera coordinate system and $T_{wo}$ in the world coordinate system. In addition, the objects' coordinates are centered around the objects and aligned with the main axis, under the assumption that all objects are placed on a horizontal plane. Therefore, it is not necessary to involve all 9 degrees of freedom (DoF) parameters. The roll and pitch angles of objects are set to zero, while the yaw angle is sampled from $-180\degree$ to $180\degree$ based on the object detector's output. The symbols used for our proposed system are illustrated in Fig. \ref{fig:coordinate_system}.

Before being transferred to the object data association process, the cuboid proposals undergo filtering and selection based on their geometric properties. We eliminate the proposals that are excessively large or distant from the camera. This selective process of cuboid proposals can enhance the robustness of the 3D reconstruction of the environment and the detection of loops.

\textit{2) Color information:}
First, we convert the RGB image into the HSV (hue, saturation, and value) color space. The areas within the bounding boxes are treated as the regions of objects in the image. We utilize the K-means++ \cite{Kmeans_plusplus} algorithm to cluster the HSV values within the regions of interest, and divide them into $K$ clusters. For each cluster, we choose its centroid as the primary color of the cluster, then represent the distribution of the cluster color as a normalized $K$-dimensional feature vector using color histogram. This feature vector can be regarded as the color information of the corresponding object. 

\begin{figure}[tp]
    \centering
    \includegraphics[width=0.45\textwidth]{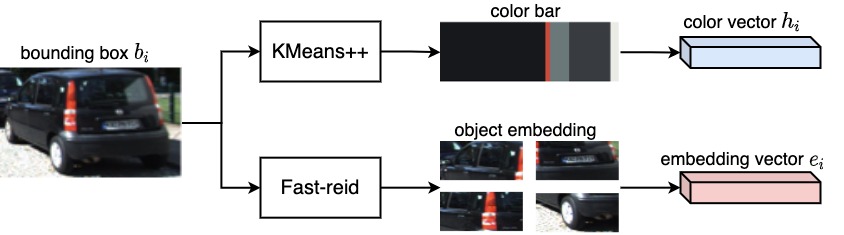}
    \caption{Illustration of the color information and embedding information extraction process for a detected object.}
    \label{fig:color_embedding}
\end{figure}

\textit{3) Embedding information:}
Similarly, we consider the regions of the bounding boxes as image patches of the objects. We employ the fast-reid method \cite{he2020fastreid} to encode the object image patches into vector representation. The fast-reid method is pre-trained in the VERI-Wild dataset \cite{lou2019large}, a large-scale vehicle ReID dataset in the wild. Fig. \ref{fig:color_embedding} depicts the color information and embedding information extraction process of detected objects.

\subsection{Object Data Association}
Data association is a crucial component of the semantic mapping model. Compared with point-based methods such as ORB-SLAM3 \cite{ORBSLAM3_TRO} and VINS-Fusion \cite{qin2019fusion}, it is easier to associate with objects in multiple frames since more information can be utilized. Suppose that there are $N$ detected objects from the object detector at frame $c_i$, which are represented as $D = \{d_i\}^N_{i=1}$. Each detection is defined by

\begin{equation*}
    d_i = \{l_i, b_i, h_i, e_i, T_{co}\},
\end{equation*}
where $l_i$ is the semantic label, $b_i$ is the bounding box, $h_i$ is the normalized vector for color histogram, $e_k$ is the object embedding from an instance-feature learning network, and $T_{co}$ is the pose of the detection in camera coordinate system. Meanwhile, there are $M$ object landmarks in the global map, represented as $\mathcal{O} = \{o_k\}^M_{k=1}$. Each object is defined by

\begin{equation*}
    o_k = \{\tilde{l}_k, \tilde{u}_k, \tilde{b}_k, H_k, E_k, T_{wo_k}\},
\end{equation*}
where $\tilde{l}_k$ is the semantic label, $H_k = \{ \tilde{h}_j\}^n_{j=1}$ and $E_k = \{ \tilde{e}_j\}^n_{j=1}$ are the sets with vectors for color histogram and object embedding from the past $n$ matches, $T_{wo_k}$ is the pose of the object in the world coordinate system, $\tilde{u}_k$ is the cuboid, and $\tilde{b}_k$ is the bounding box.

The bounding box $\tilde{b}_k$ is predicted by $\tilde{u}_k$. Specifically, we project the eight corners of $\tilde{u}_k$ onto the image and obtain the minimum and maximum $x$-$y$ coordinates of the projected pixels to form a rectangle \cite{cubeslam} as

\begin{equation}
\begin{aligned}
    p_{lk}{}={}& \mathrm{min}\{\pi(\mathbf{R}[\pm d_x, \pm d_y, \pm d_z]/2 + \mathbf{t})\}, \\
    p_{rk}{}={}& \mathrm{max}\{\pi(\mathbf{R}[\pm d_x, \pm d_y, \pm d_z]/2 + \mathbf{t})\},
\end{aligned}
\end{equation}
where $\pi(\cdot)$ represents the camera projection function, while $\mathbf{R}$ and $\mathbf{t}$ denote the rotation and translation matrix of the pose of the frame $c_i$, respectively. 

The object data association algorithm aims to match objects in the object map with the detected ones as accurately as possible by computing their similarity

\begin{equation}
    \mathbf{W}_d(i,k) = 
    \begin{cases}
    s_d(d_i,o_k), &{\text{if}}\ {l_i = \tilde{l}_k},  \\
    \mathbf{0},                    &{\text{otherwise}},
    \end{cases}
\end{equation}
where the matching matrix $\mathbf{W}_d(i,k)$ between a detected object $d_i$ and an object $o_k$ in the global map is computed based on bounding boxes, color histogram, and object embedding as

\begin{equation}
    \begin{aligned}
        s_d(d_i,o_k) &= \lambda {iou}(b_i,\tilde{b}_k) +  \lambda(1-\lambda) his(h_i, H_k) \\ 
        &+ (1-\lambda)^2 dis(e_i, E_k),
    \end{aligned}
\end{equation}
where
\begin{align}
     iou(b_i,\tilde{b}_k) &= \frac{b_i \cap \tilde{b}_k}{b_i \cup \tilde{b}_k}, \\
     his(h_i, H_k) &= \frac{1}{\| H_k\|}\sum_{\tilde{h}_j \in H_k} h_i \tilde{h}_j, \\
     dis(e_i, E_k) &= \frac{1}{\| E_k\|}\sum_{\tilde{e}_j \in E_k} e_i \tilde{e}_j,
\end{align}
$iou(\cdot)$ is calculated between the bounding box $b_i$ and the predicted bounding box $\tilde{b}_k$. The metric distance $his(\cdot)$ of the histogram vector is calculated between $h_i$ and each vector $\tilde{h}_j$ in $H_k$. The metric distance $dis(\cdot)$ of embedding is calculated between $e_i$ and each embedding $\tilde{e}_j$ in $E_k$. Additionally, a hyperparameter $\lambda$ is used to balance between the $iou(\cdot)$, $his(\cdot)$ and $dis(\cdot)$ terms.

\subsection{Object Pose Refinement}\label{sec:semantic mapping}
When the objects are successfully associated, we utilize the camera motion model and object spatial location to refine the objects' poses. Specifically, we employ the bundle adjustment approach \cite{ba} to jointly optimize the poses of the cameras and objects, which formulates the optimization as a nonlinear least squares optimization problem

\begin{equation}\label{equ:2}
    T_{wo}^*, T_{wc}^*= \arg\min\limits_{T_{wo}, T_{wc}}{\sum_{c=c_0}^{c_n} \| r(T_{wo}, T_{wc}) \|^2},
\end{equation}
where
\begin{equation}\label{equ:3}
    r(T_{wo}, T_{wc}) = \log( T_{wo}^{-1} T_{wc} T_{co} )^\vee_{\mathfrak{se}3}.
\end{equation}

The indices from $c_0$ to $c_n$ indicate the frames where the object is tracked. $r(\cdot)$ represents the spatial measurement errors between the object and the matched detection. Specifically, $T_{wc}$, $T_{wo}$, and $T_{co}$ represent the poses of the current camera, the object, and the matched detection, respectively. $T_{co}$ is in the camera coordinate system. The $\log(\cdot)$ operation is applied to map the errors in the $SE(3)$ space into a 6 DoF target vector space, i.e., $e(\cdot) \in \mathbb{R}^6$. To jointly adjust the object poses and camera poses, we adopt the sliding window optimization mechanism based on the approach in VINS-Fusion \cite{qin2019fusion}. In our implementation, the objects that have been tracked more than 4 times will be optimized. The optimization problem is solved by the Gauss-Newton algorithm \cite{GN} from the Ceres library\cite{ceres-solver}.

Finally, we can construct an object-level semantic map with more accurate object poses in the environment, as shown in Fig. \ref{fig:semiMap}.

\begin{figure}[t]
    \centering
    \includegraphics[width=0.45\textwidth]{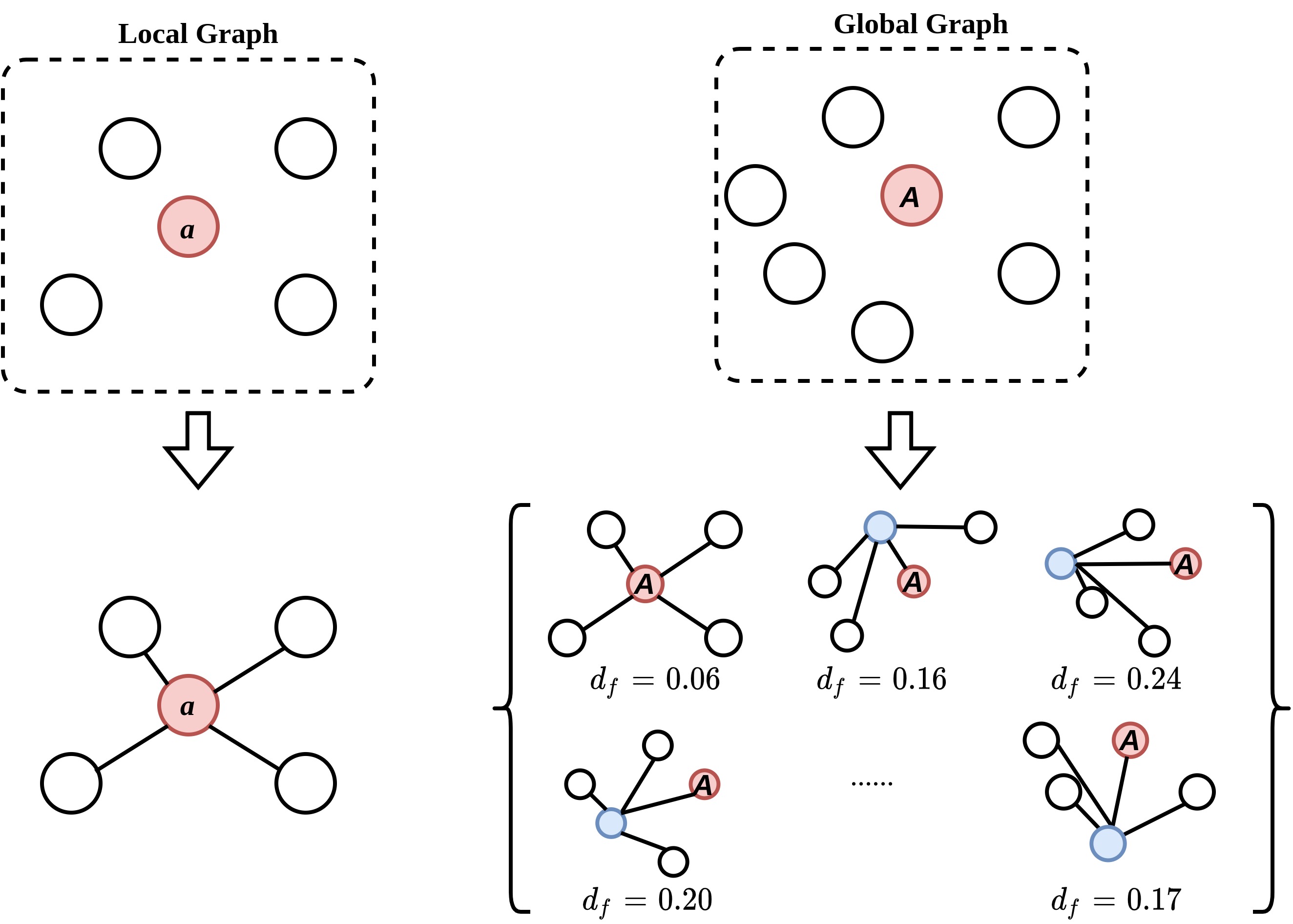}
    \caption{Illustration of the graph extraction and the graph matching based on the spatial layout of objects. Edges in the graph represent the spatial distance between objects. Graph matching aims to find correspondence objects between the local and global maps based on the similarity of graphs.}
    \label{fig:LCD1}
\end{figure}
\section{Loop Closure With Semantic Landmarks} \label{sec:loop_closure_with_semantic_landmarks} 
Firstly, we represent the objects in the local and global maps as vertices and set the position of the center of the vertices in the world coordinate system. We construct the relative relationships between vertices and their adjacent vertices. Then, we perform two-stage topology graph matching to identify the correspondences between the local and global graphs based on the consistency of spatial layout and vertex properties. In the initial stage, we build neighbor topological structures and select candidate matches based on their spatial layout consistency in the local and global maps. In the subsequent stage, we compute semantic similarity and select the most similar match as the corresponding object. Finally, we mitigate the drift errors and correct the camera poses by aligning the matched objects.

\subsection{Graph Generation}
The associated and refined objects in the map form a graph $G=(V, E)$, where each vertex $v_i \in V$ contains the center position, semantic label $l_i$, color histogram vector $h_i$, and object embedding $e_i$. Each edge $e_{edge}(i_1,i_2) \in E$ is determined by the spatial distance between the centers of object $o_{i_1}$ and object $o_{i_2}$ as

\begin{equation}
   e_{edge}(i_1,i_2) = \|t( T_{wo_{i_1}}^{-1} T_{wo_{i_2}} )\|_2,
\end{equation}
where the function $t(\cdot)$ calculates the translation vector of the poses.

To preserve the graphs' topology, we adopt a proximity strategy where each vertex is connected to the $K$ nearest neighbors based on their spatial distance. In this way, we can obtain two graphs $G_l$,  $G_g$ from the local and global maps, respectively.

\subsection{Spatial Layout Matching}\label{sec:loop_detection}
Given the local graph $G_l=(V_l, E_l)$ with $M$ vertices and the global graph $G_g=(V_g, E_g)$ with $N$ vertices, we seek a collection of putative matches as
\begin{equation*}
    \mathcal{S}=\{(v_{\mathcal{L}_i}, v_{\mathcal{G}_j})\}_{
    \substack{i=1, \dots, M \\ j=1, \dots, N}
    },
\end{equation*}
where $v_{\mathcal{L}_i} \in G_l$ and $v_{\mathcal{G}_j} \in G_g$. There are $M\times N$ possible matching pairs in total. The matching stage aims to identify multiple feasible correspondences between graphs, which fit the vertex's spatial layout. For a potential pair $(v_{\mathcal{L}_i}, v_{\mathcal{G}_j})$, we find the $K$ nearest neighbors of $v_{\mathcal{L}_i}$ in $G_l$ and $v_{\mathcal{G}_j}$ in $G_g$, and represent the neighbors as topological structures with weight vectors $\mathbf{w}_{\mathcal{L}_i}$ and $\mathbf{w}_{\mathcal{G}_j}$, respectively. The size of the vectors is $1\times K$ and the vectors are sorted and normalized, such that the sum of their elements is equal to one. The spatial layout difference of the potential pair can be determined as 
\begin{equation}
    d_f(v_{\mathcal{L}_i}, v_{\mathcal{G}_j}) = \|\mathbf{w}_{\mathcal{L}_i} - \mathbf{w}_{\mathcal{G}_{j}} \|_2.
\end{equation}

The matched objects would have lower difference in their neighborhoods, as illustrated in Fig. \ref{fig:LCD1}. Therefore, we accept the object pairs with difference less than a threshold $\delta$ as candidate matching.

\subsection{Semantic Properties Verifying}
Limiting the total number of correspondences is necessary as it can reduce the possibility of false matching. It is achieved through filtering based on semantic properties consistency, such as their geometry properties (semantic labels, object scales, and poses) and embedding properties (colors and embedding vectors). Following the spatial layout matching stage, a set of candidate matches can be obtained and denoted as $\mathcal{C} = \{(v_i, \tilde{v}_{i'})\}^C_{i=1}$, where $\tilde{v}_{i'} \in G_g$ is the matched object of $v_i \in G_l$. The consistency of their semantic properties is calculated as

\begin{equation}
    \mathbf{W}_l(i,i') = 
    \begin{cases}
    s_l(v_i, \tilde{v}_{i'}),   &{\text{if}}\ {l_i = \tilde{l}_{i'}},  \\
    \mathbf{0},                       &{\text{otherwise}},
    \end{cases}
\end{equation}
where
\begin{align}
& s_l(v_i, \tilde{v}_{i'}) = \mu(\mathbf{d}_s + \mathbf{d}_p) + (1-\mu)(\mathbf{d}_c + \mathbf{d}_e), \\
& \mathbf{d}_s = \exp(- \| [d_{xi}, d_{yi}, d_{zi}] - [\tilde{d}_{xi'}, \tilde{d}_{yi'}, \tilde{d}_{zi'}]\|_2), \\
& \mathbf{d}_p = \exp(- \| \mathbf{t}_i - \tilde{\mathbf{t}}_{i'} \|_2), \\
& \mathbf{d}_c = \exp(- \| h_i  \tilde{h}_{i'} \|_2), \\
& \mathbf{d}_e = \exp(- \| e_i  \tilde{e}_{i'} \|_2).
\end{align}
$l$ refers to the semantic label of the vertex. The $[d_x,d_y,d_z]$ and $\mathbf{t}$ represent the scale and position as the geometry properties of the vertex. Additionally, $h$ and $e$ denote the color and object vector as embedding properties of the vertex. A hyperparameter $\mu$ is incorporated to balance between the impact of geometry and embedding properties. Ultimately, we consider the matched pairs with similarity $s_l(v_i, \tilde{v}_{i'})$ greater than a threshold $\tau$, i.e., $s_l(v_i, \tilde{v}_{i'}) \ge \tau$, as the final matches.

\begin{figure}[t]
    \centering
    \includegraphics[width=0.45\textwidth]{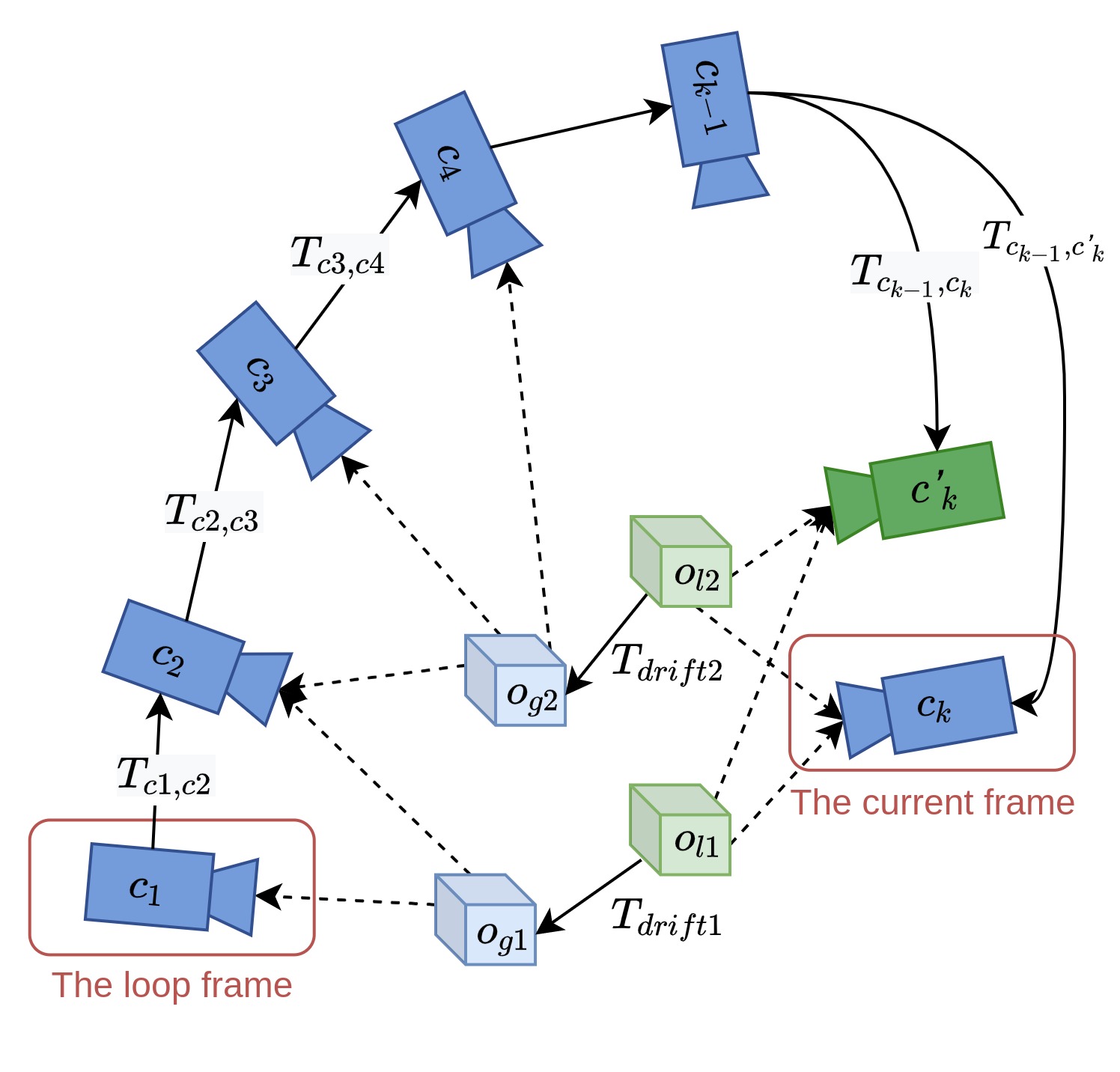}
    \caption{Illustration of object-level pose graph with camera poses and object poses. }
    \label{fig:pg}
\end{figure}

\subsection{Pose Graph Optimization}
When detected objects have multiple matches through the object-level loop detection method, the loop frame is determined to be the first frame in which any object in the global map is detected. Fig. \ref{fig:pg} depicts the current frame as $c_k$, and the objects $o_{l1}$ and $o_{l2}$ matching with $o_{g1}$ and $o_{g2}$, respectively. In this situation, we identify the first frame from all the frames observed for $o_{g1}$ and $o_{g2}$, which is $c_1$, as the loop frame associated with $c_k$.

With a loop frame associated with the current frame, we can correct the camera pose drift error by aligning the matched objects in the local and global maps. The matched objects in the maps are represented as $\mathcal{O}_l=\{o_{li}\}^N_{i=1}$ and $\mathcal{O}_g=\{o_{gi}\}^N_{i=1}$, respectively. The transformation $T_{drift}$ for correcting the drift can be computed as

\begin{equation}
    T_{drfit}^* =  \arg\min\limits_{T_{drift}} {\sum_{i=1}^{N} \|r_{oo}(T_{wo_{li}}, T_{wo_{gi}})) \|_{\Sigma_{li,gi}}},
\end{equation}
where
\begin{equation}
    r_{oo}(T_{wo_{li}}, T_{wo_{gi}}) = \log(T_{drift}^{-1} T_{wo_{li}}^{-1} T_{wo_{gi}})^\vee_{\mathfrak{se}3}.
\end{equation}
$r_{oo}(\cdot)$ is the measurement error for the object-object constraint. $\| r\|_{\Sigma} = r^T\Sigma^{-1}r$ denotes the Mahalanobis distance. To solve this nonlinear least squares problem, we also employ the Gauss-Newton algorithm \cite{GN} provided in the Ceres library \cite{ceres-solver}.

Furthermore, we can correct the camera pose $T_{wc_k}$ of the current frame $c_k$ using the drift correct transformation as

\begin{equation}
    T_{wc'_k} = T^{-1}_{drift}T_{wc_k}.
\end{equation}

When the pose of the current frame is corrected, we proceed to perform frame pose graph optimization to adjust the poses of frames between the loop and current frames. To be more specific, we denote the poses of these frames as $\mathcal{X} = \{ T_{wc_i}\}^M_{i=1}$, which is optimized by

\begin{equation}
    \mathcal{X} = \arg\min\limits_{\mathcal{X}} {\sum_{i=1}^{M-1} \|r_{cc}(T_{wc_i}, T_{wc_{i+1}})) \|_{\Sigma_{i,i+1}}},
\end{equation}
where 
\begin{equation}
    r_{cc}(T_{wc_i}, T_{wc_{i+1}}) = \log(T_{c_i,c_{i+1}}^{-1} T_{wc_i}^{-1} T_{wc_{i+1}})^\vee_{\mathfrak{se}3}.
\end{equation}
$r_{cc}(\cdot)$ is the measurement error for the camera-camera constraint. After the optimization, we synchronize the object poses and the entire trajectory with the camera motion.

\begin{figure*}[t]
    \centering
    \includegraphics[width=0.95\textwidth]{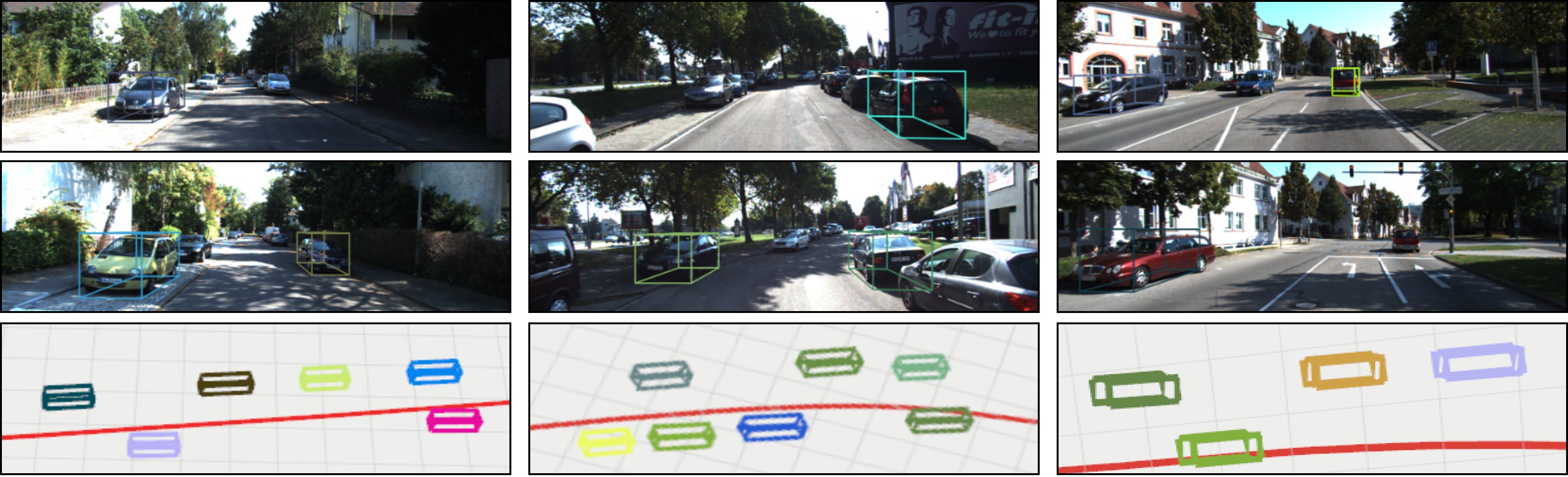}
    \caption{Qualitative results of our semantic mapping. We project the estimated 3D bounding box on two sequential images and semantic maps. Note that the red lines represent the relative trajectory with the ego-camera poses at each time.}
    \label{fig:semiMap}
\end{figure*}

\begin{table*}[ht]\caption{3D Object Detection Results with 3D IoU metric on KITTI Benchmark}\label{table:1}
\centering
\renewcommand{\arraystretch}{1.2}
\begin{tabular}{l|ccc|ccccc|c}
\hline
\multicolumn{1}{c|}{\multirow{2}{*}{KITTI Benchmark}} &
  \multicolumn{3}{c|}{Tracking Sequence} &
  \multicolumn{5}{c|}{Raw Sequence} &
  \multirow{2}{*}{Mean} \\ \cline{2-9}
\multicolumn{1}{c|}{}                   & 07    & 09    & 11    & 01    & 23    & 35    & 39    & 61    &       \\ \hline
CubeSLAM\cite{cubeslam}                 & 0.210 & 0.190 & 0.190 & 0.576 & 0.350 & 0.523 & 0.590 & 0.500 & 0.394 \\
VisualDet3D\cite{liu_ground-aware_2021}   & 0.455 & 0.287 & 0.348 & 0.578 & 0.320 & 0.555 & 0.534 & 0.540 & 0.454 \\
Ours &
  \textbf{0.586} &
  \textbf{0.311} &
  \textbf{0.446} &
  \textbf{0.605} &
  \underline{0.347} &
  \textbf{0.572} &
  \underline{0.577} &
  \textbf{0.610} &
  \textbf{0.512} \\ 
  \hline
\end{tabular}
\end{table*}

\begin{table}[t]\caption{details of sensors used for collecting the HALV dataset}\label{table: HALV_sensor}
\centering
\renewcommand\arraystretch{1.2}
\resizebox{0.48\textwidth}{!}{
\begin{tabular}{c|c|l}
\hline
Sensor &
  Rate &
  \multicolumn{1}{c}{Characteristics} \\ \hline
Camera &
  30Hz &
  \begin{tabular}[c]{@{}l@{}}1280 x 720 Resolution (used in our experiments)\\ 24-bit RGB\\ Field of View: 110°(H) x 70°(V) x 120°(D) max\\ Shutter Sync: Electronic Synchronized Rolling Shutter\end{tabular} \\ \hline
IMU &
  400Hz &
  \begin{tabular}[c]{@{}l@{}}Accelerometer\\ Gyroscope \end{tabular} \\ \hline
\end{tabular}
}
\end{table}

\section{Experimental Results}
\label{sec:results}
The proposed system is evaluated on both the KITTI dataset \cite{inproceedingskitti} and our HALV dataset. All the experiments are conducted on a computer with an Intel i7-9700 CPU operating at 3.00GHz and 16GB of RAM. In the first experiment (see Section \ref{exp: mapping}), the proposed semantic mapping system is compared with representative algorithms on the KITTI dataset, and the evaluation metric is the intersection over union (IoU). The second (see Section \ref{exp: detection}) and third (see Section \ref{exp: localization}) experiments are carried out on our collected HALV dataset to evaluate the loop closure performance in significant viewpoint-changing scenes, using the absolute trajectory errors (ATE) as the evaluation metric. The system's robustness and accuracy are evaluated through testing on several cases within the HALV dataset. Part of the code is publicly available on \url{https://github.com/jixingwu/SS-LCD.git}

\subsection{Dataset} 
\textit{1) KITTI dataset:} It is a popular benchmark dataset for computer vision research in autonomous driving. It comprises a diverse set of sensor data collected from a mobile platform in urban and highway environments, including monocular/stereo cameras, LIDAR, and GPS/IMU. The dataset contains various tasks, such as object detection, tracking, segmentation, and depth estimation, with high-quality annotations for each task. Among them, the tracking and raw sequences have been widely used for evaluating and developing algorithms related to 3D perception. Therefore, for the first experiment, we have chosen three KITTI tracking sequences labeled "00xx" and five KITTI raw sequences labeled "2011$\_$0926$\_$00xx" that contain the most static objects and ground truth annotations.

\begin{figure}[t]
    \centering
    \includegraphics[width=0.48\textwidth]{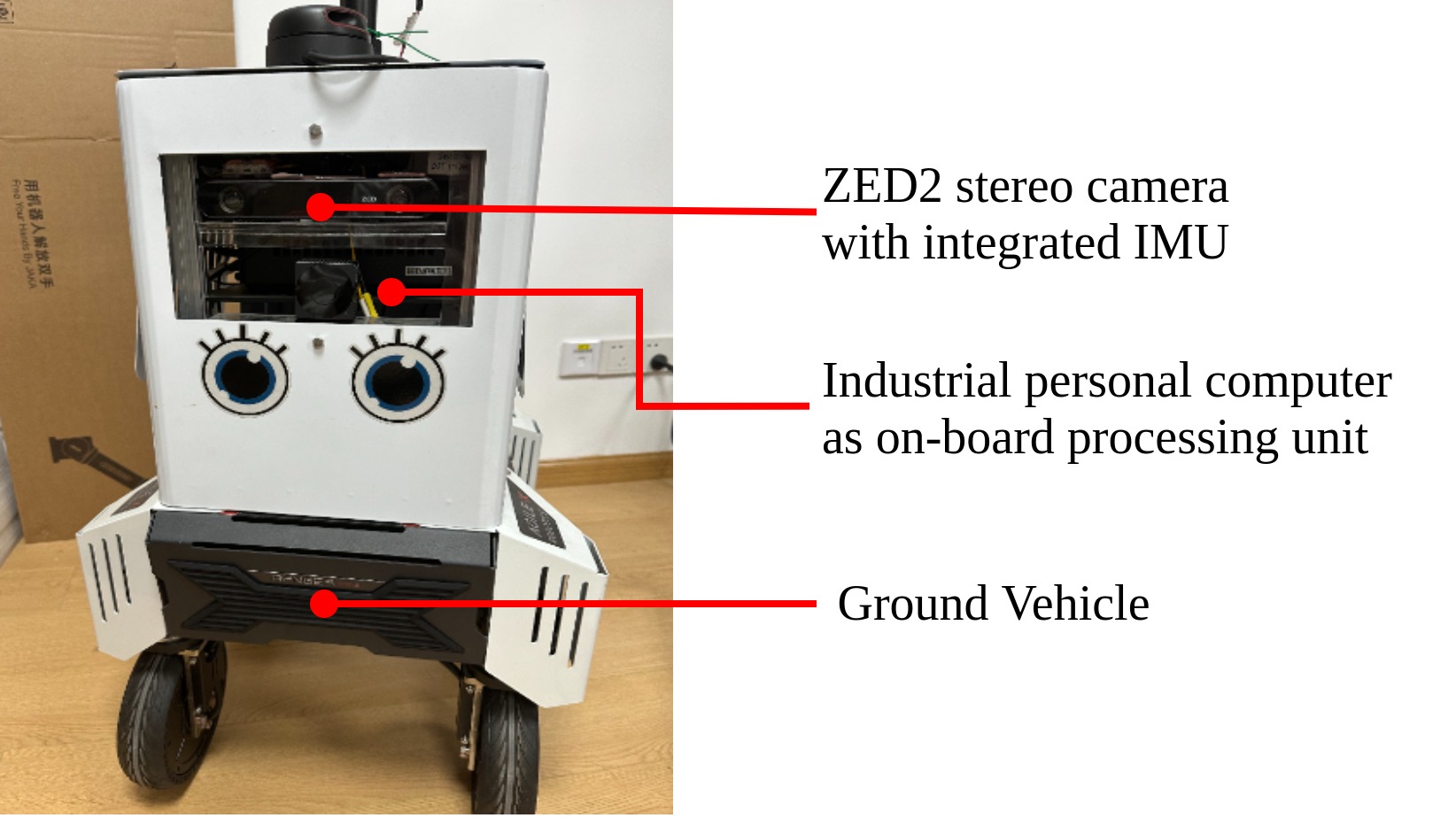}
    \caption{Our data collection platform.}
    \label{fig:robot}
\end{figure}

\textit{2) HALV dataset:} It is our collected dataset for loop closure research with large viewpoint changes in an outdoor parking environment. We have employed a ground vehicle with a ZED2 stereo camera and a 9-axis IMU module to collect four challenging sequences covering various observation distances and viewpoints. The characteristics of sensors are summarized in Table \ref{table: HALV_sensor} and the data collection platform is shown in Fig. \ref{fig:robot}. For recovering the complete trajectories of the ground vehicle, we employed VINS-Fusion \cite{qin2019fusion} and designed trajectories that enable to detect loop closures and reduce drift errors, making it a reliable source of ground truth for the trajectories. We then segmented the data from the four image sequences and removed the end to create a loop closure scene with significant viewpoint changes. In HALV, the four trajectories with three different shapes are collected: rectangles, straight lines, and curves. Our dataset is publicly available on \url{https://1drv.ms/u/s!Al-c847UhpgZbFsNYTaiLMDZx-M?e=En21bq}

\subsection{Semantic Mapping Performance} 
\label{exp: mapping}
After obtaining the raw 3D detection results of objects using VisualDet3D \cite{liu_ground-aware_2021}, we consider both the spatial consistency and temporal association to optimize the poses of these objects. It should be noted that the accuracy of 3D object detection can affect the performance of the semantic mapping process. 

Table \ref{table:1} shows quantitative evaluation results of 3D object detection on the KITTI dataset. We evaluate the 3D object IoU on the KITTI dataset and compare our results with VisualDet3D \cite{liu_ground-aware_2021}, and CubeSLAM \cite{cubeslam}. CubeSLAM introduces a 3D object detection approach based on 2D bounding box regression and an object-based pose estimation method. The world frame is built on the horizontal plane, and the objects' roll/pitch angles are set to zero, which is consistent with our assumption. In our implementation, we only evaluate the objects with a detection probability greater than 0.8, as reported in the object SLAM experiments in the CubeSLAM work \cite{cubeslam}. The performance of our method decreases slightly on two KITTI raw sequences due to relative pose drift and image sequence shortening. Note that the accuracy of VisualDet3D also affects the performance of our method. 

Fig. \ref{fig:semiMap} shows the qualitative results of our proposed object-level semantic mapping. With precise 3D cuboids, we project them onto two consecutive images. The third row in the figure represents the bird's view map, where different cuboids are rendered in cars' colors, and red lines indicate the camera trajectory.

\begin{figure}[t]
    \centering
    \includegraphics[width=0.48\textwidth]{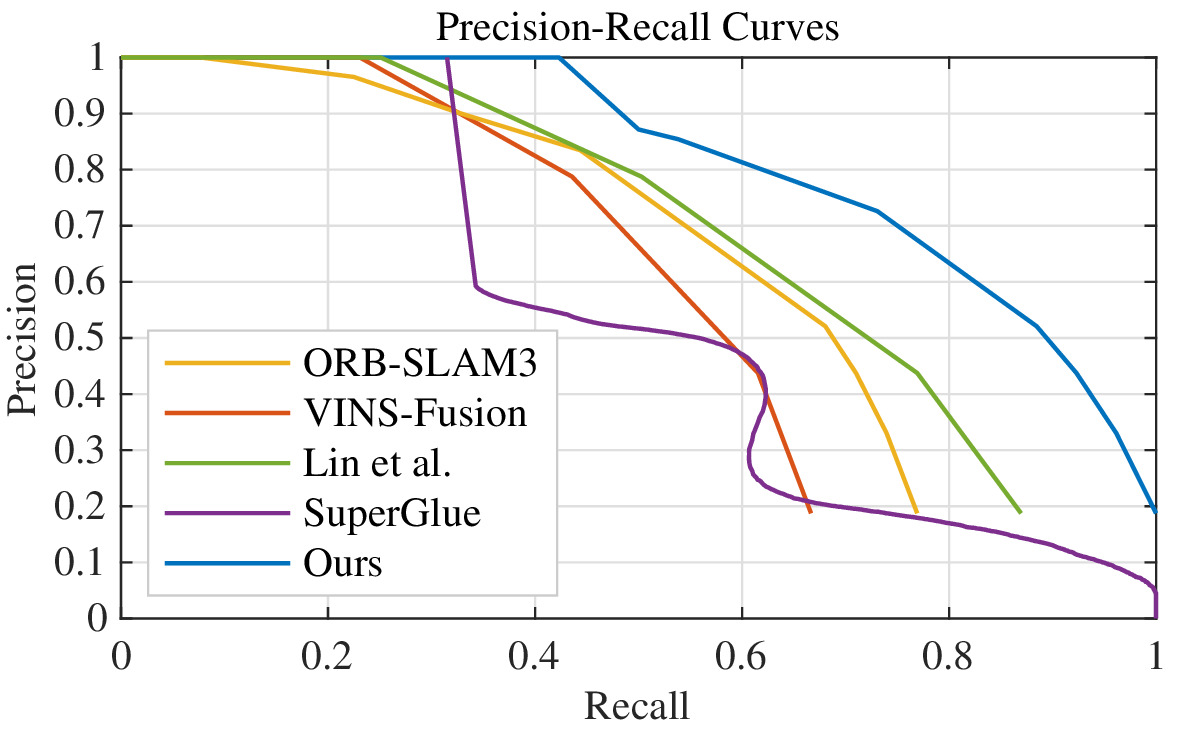}
    \caption{The precision-recall curves on loop detection.}
    \label{fig:PR}
\end{figure}

\begin{figure}[t]
    \centering
    \subfigure[distance error in camera pose estimation]{
    \includegraphics[width=0.4\textwidth]{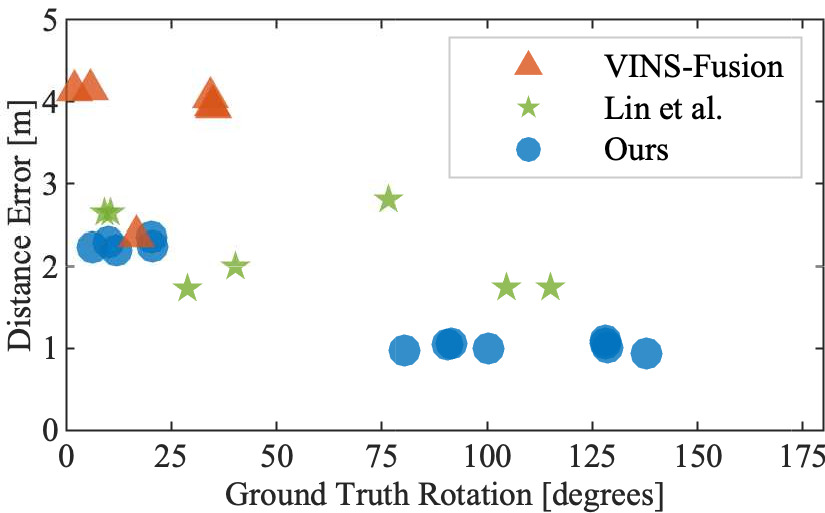}
    }
    
    \subfigure[rotation error in camera pose estimation]{
    \includegraphics[width=0.4\textwidth]{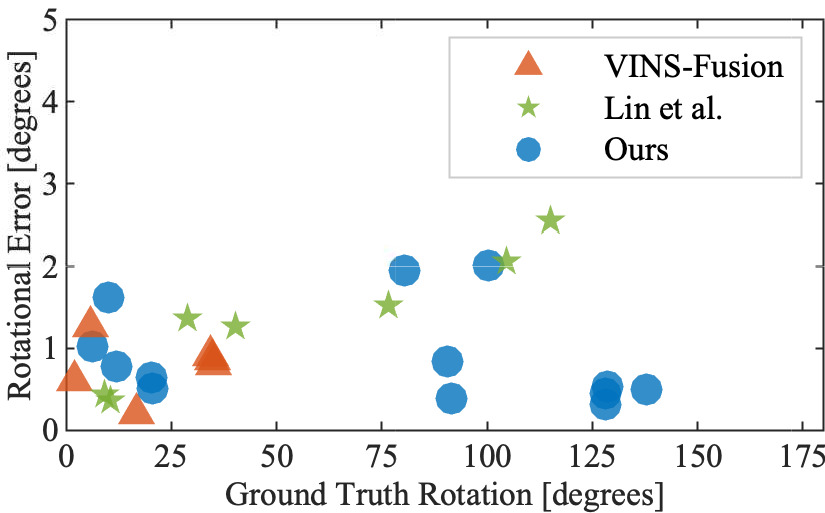}
    }
    \caption{We show the camera pose estimation error after the loop closure correction. Our method effectively reduces the camera drift error compared to the approach proposed by Lin et al., even in cases with viewpoint changes over 125 degrees. In contrast, VINS-Fusion approach is limited to less than 50 degrees.}
    \label{fig:dis_and_rot}
\end{figure}

\begin{figure}[t]
    \centering
    \includegraphics[width=0.4\textwidth]{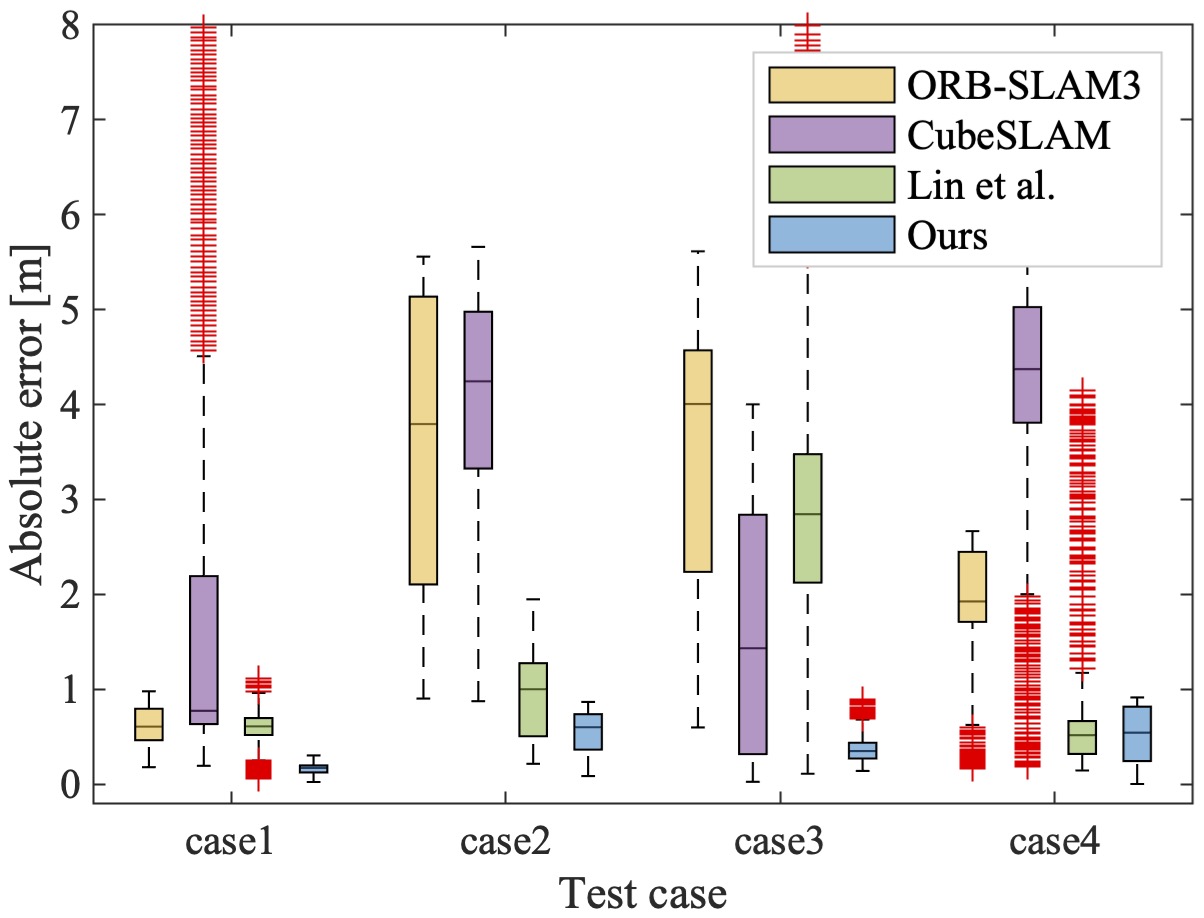}
    \caption{Statistical absolute translation error comparison on the HALV dataset.}
    \label{fig:box}
\end{figure}

\begin{figure*}[htp]
    \centering
    \subfigure[Trajectory and localization error comparison on case 4]{
    \includegraphics[width=0.54\textwidth]{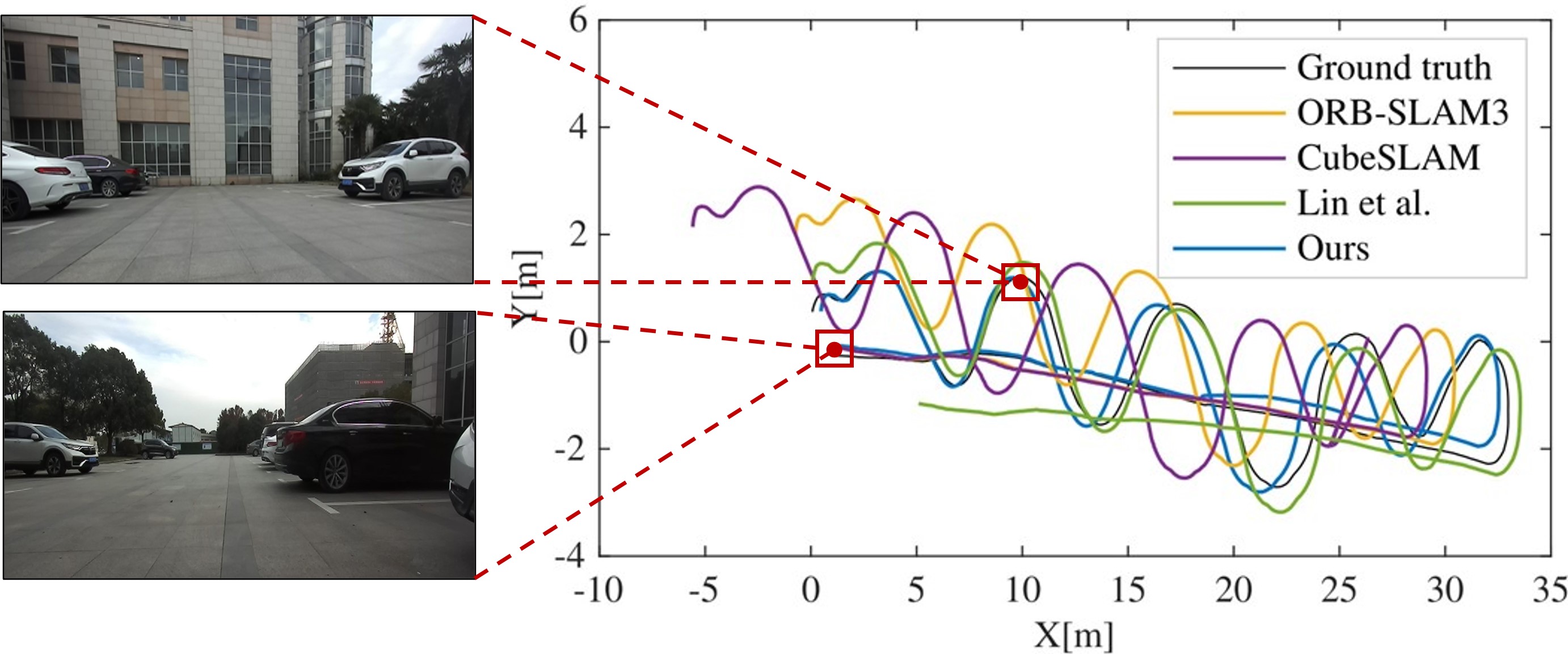}
    \includegraphics[width=0.37\textwidth]{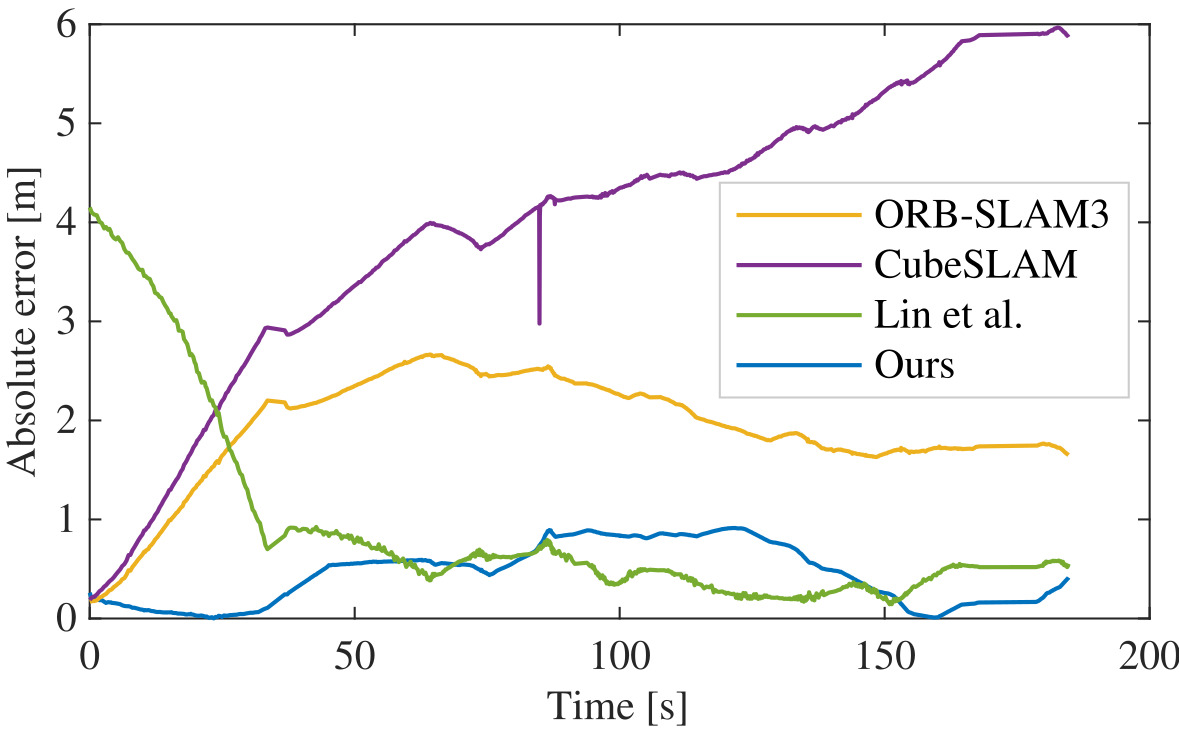}
    }
    
    \subfigure[Trajectory and localization error comparison on case 2]{
    \includegraphics[width=0.54\textwidth]{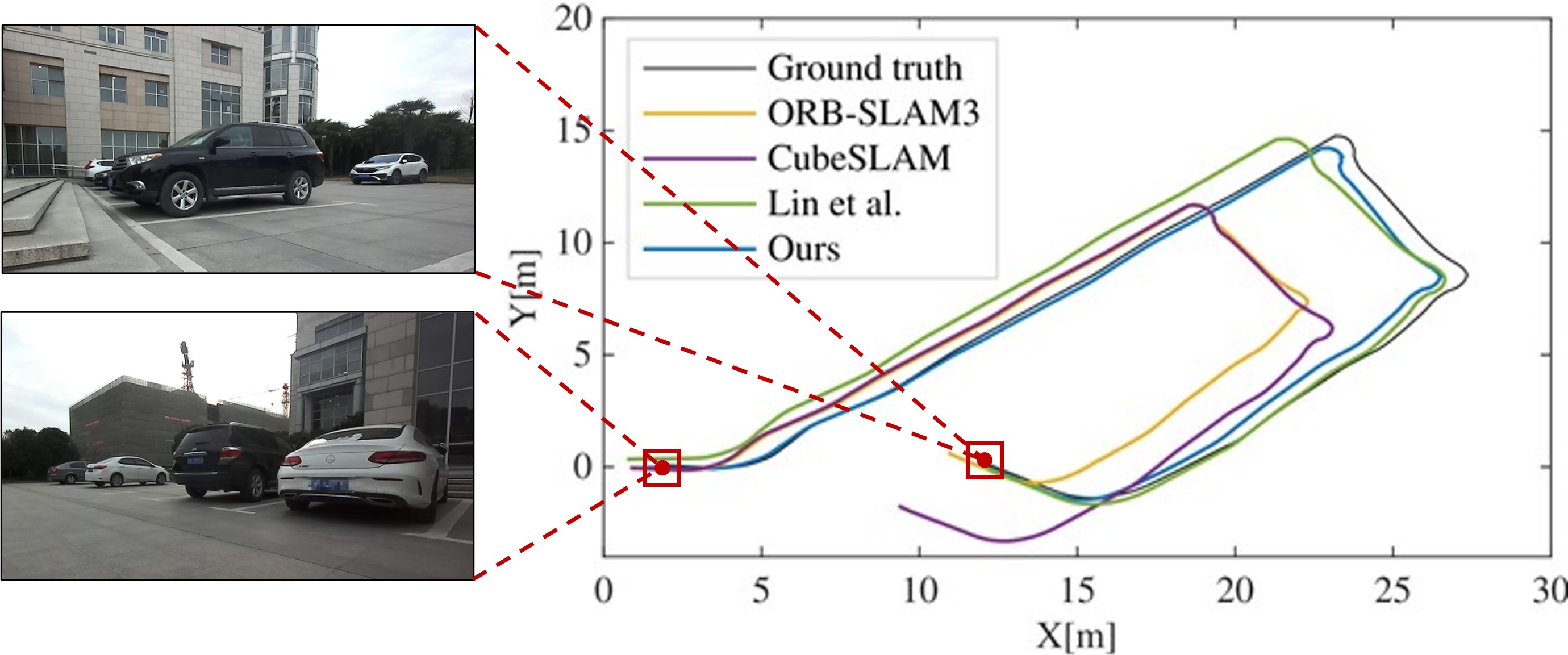}
    \includegraphics[width=0.37\textwidth]{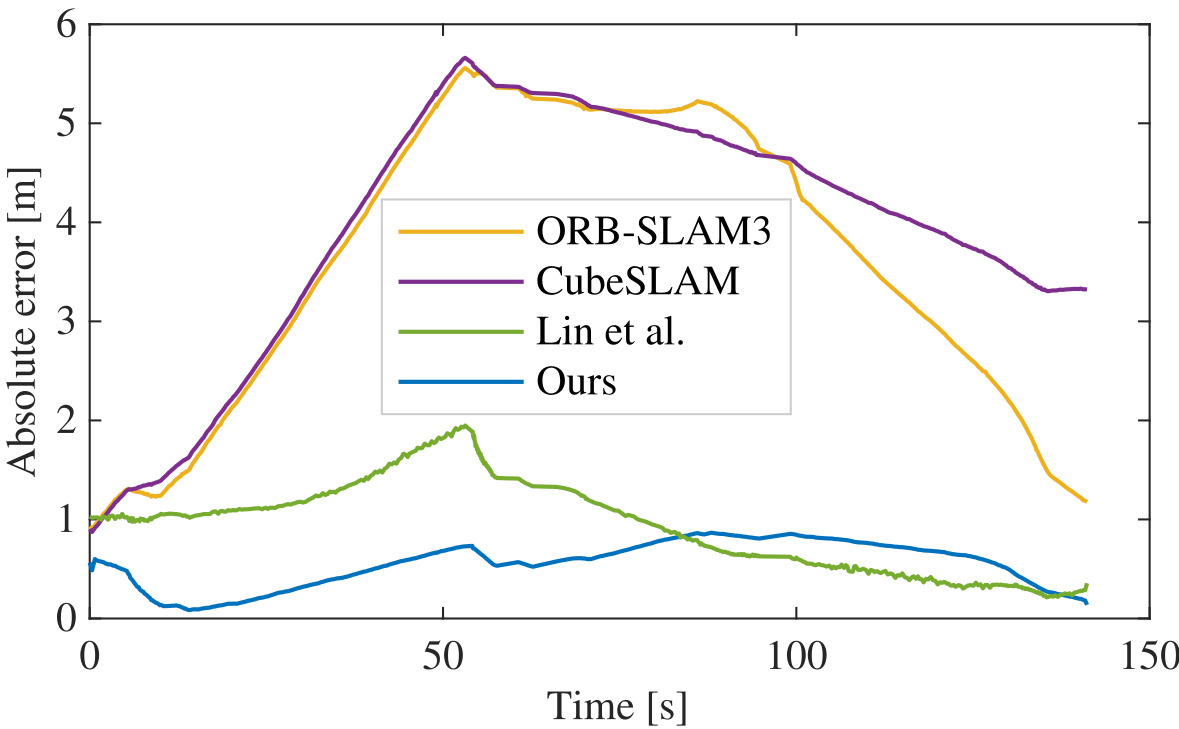}
    }
    
    \subfigure[Trajectory and localization error comparison on case 3]{
    \includegraphics[width=0.54\textwidth]{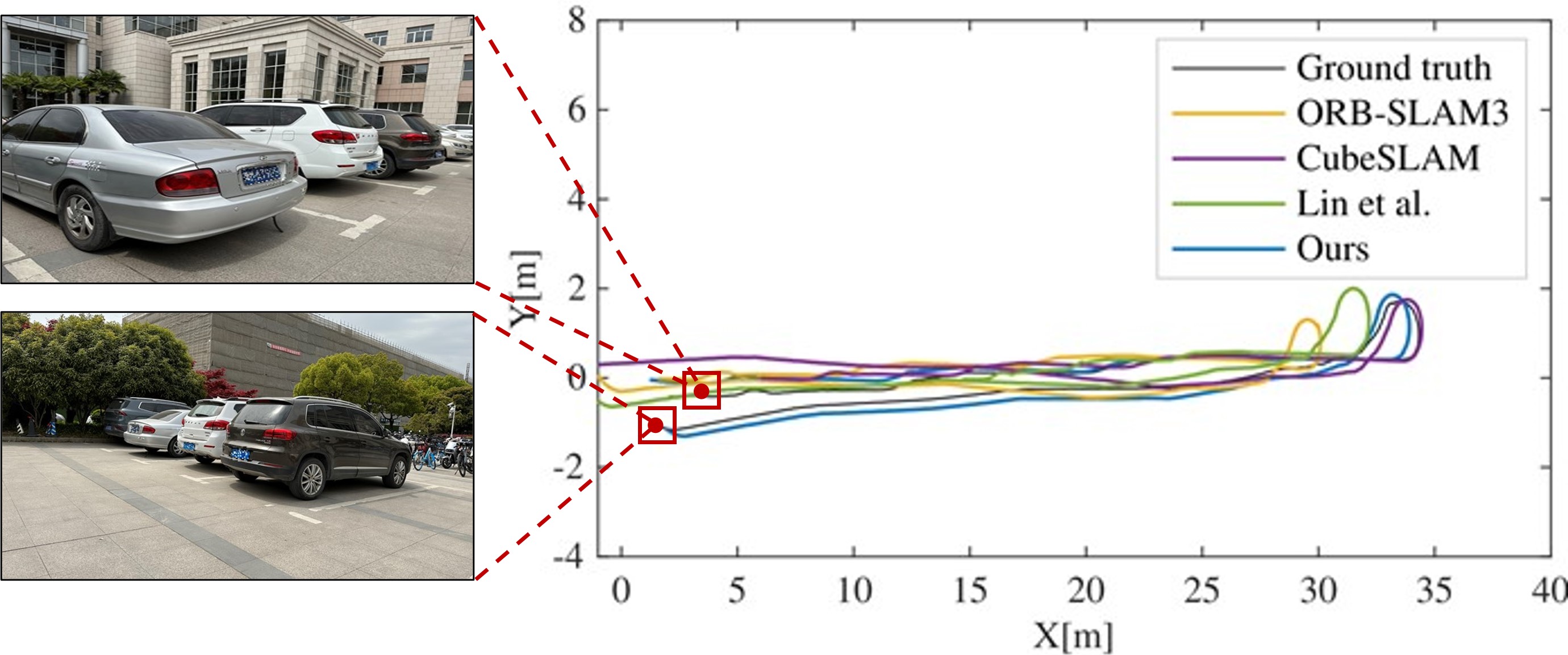}
    \includegraphics[width=0.37\textwidth]{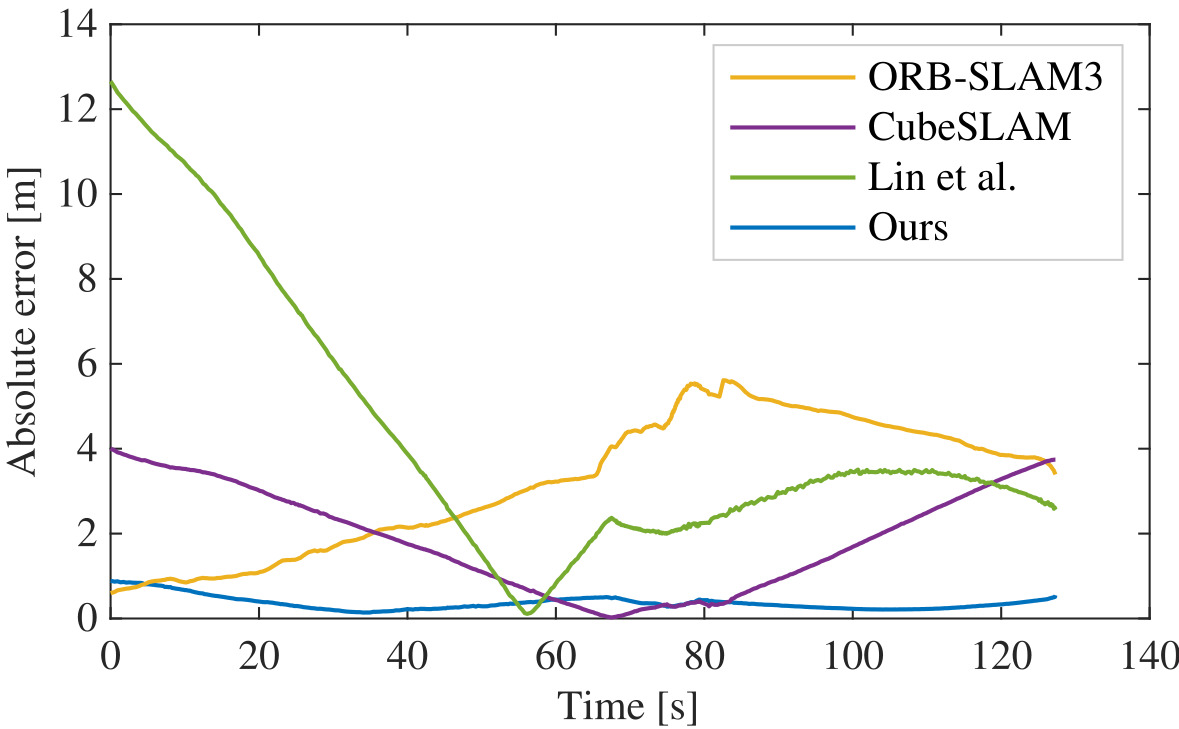}
    }
    
    \subfigure[ Trajectory and localization error comparison on case 1]{
    \includegraphics[width=0.54\textwidth]{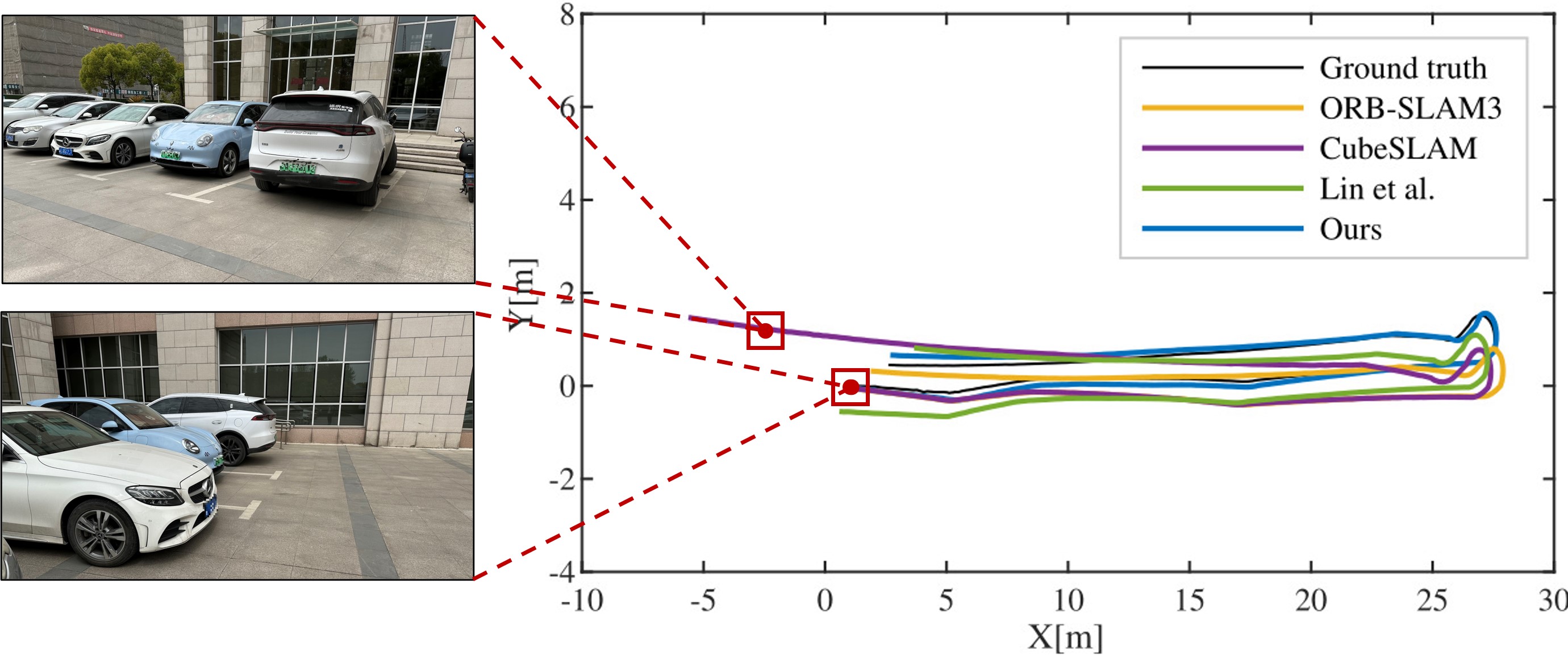}
    \includegraphics[width=0.37\textwidth]{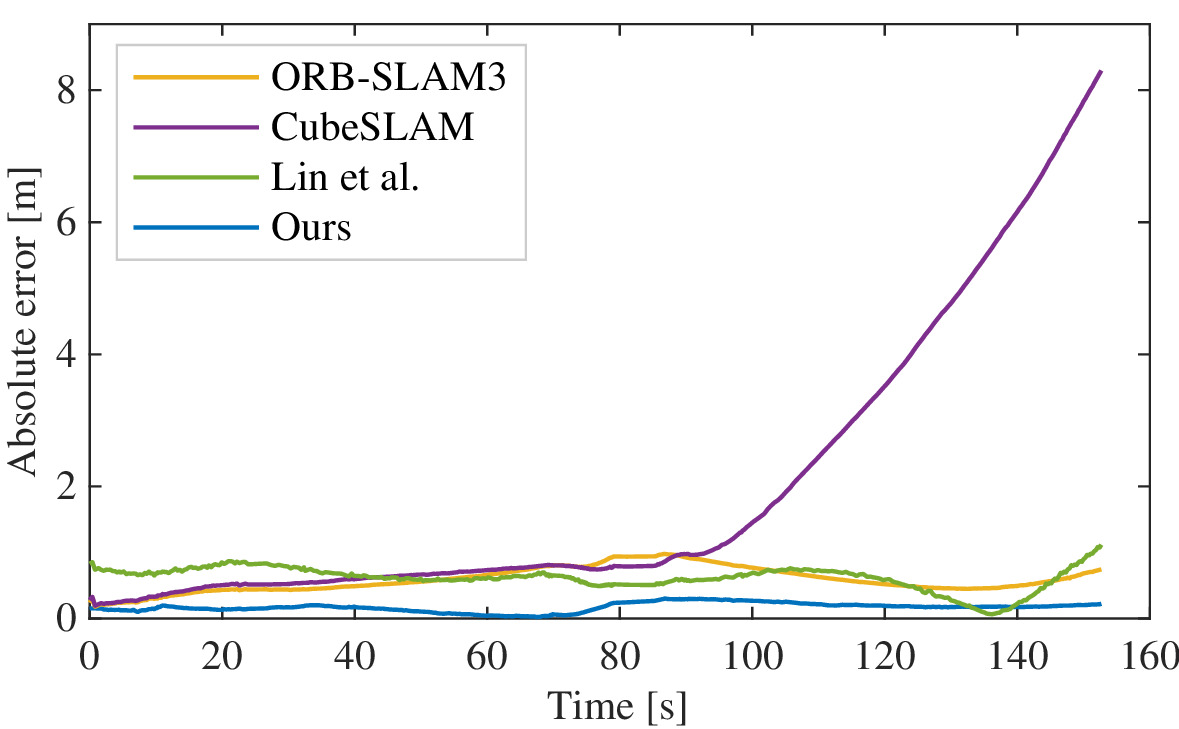}
    }
    
    \caption{Real world tests collected by ZED 2 and result comparisons with ORB-SLAM3 \cite{ORBSLAM3_TRO}, CubeSLAM \cite{cubeslam} and Lin et al. \cite{lin_topology_2021}.}
    \label{fig:traj}
\end{figure*}

\subsection{Loop Detection Performance} 
\label{exp: detection}
In this section, the proposed system is benchmarked and compared with the most relevant state-of-the-art SLAM systems in a few large viewpoint scenes from the HALV dataset.

Fig. \ref{fig:PR} shows precision-recall (PR) curves of different methods on the HALV dataset. ORB-SLAM3 and VINS-Fusion utilize DBoW2 \cite{DBOW2} for feature retrieval and loop closure detection. DBoW2 is constructed based on the BRIEF binary descriptors \cite{brief} extracted from corner detectors.
In contrast, Lin et al. \cite{lin_topology_2021} propose object-based 3D graph construction and edit distance minimization-based loop detection. We have reproduced the implementation of this method ourselves since the code has not been released. To ensure a fair comparison, the method's object extraction and pose estimation are performed by VisualDet3D \cite{liu_ground-aware_2021} and VINS-Fusion \cite{qin2019fusion}, respectively, the same as that in our method. 

SuperGlue \cite{superglue} is an excellent deep-learning method for place recognition and feature matching. It incorporates a versatile attention-based contextual aggregation mechanism that can reason about the underlying 3D scene and assign features simultaneously. According to the work \cite{garg_look_2019}, the score $S$ of the matched images for SuperGlue is computed as

\begin{equation}
    S(I_i, I_j) =
    \begin{cases}
    \frac{1}{N}\sum_{n=0}^{N} s(P_n),   &{\text{if}}\ {N \ge 5},  \\
    0,                                  &{\text{otherwise}},
    \end{cases}
\end{equation}
where $(I_i, I_j)$ denotes a pair of matched images, $s(P_n)$ denotes the normalized confidence of a point match, and $N$ denotes the number of point matches in the image pair.

The PR curves are generated based on the localization threshold $\tau_L$, which represents the maximum allowed distance between the estimated robot position $t_c$ and the ground truth position $t_{gt}$. A loop closure is considered to be true if the distance between $t_c$ and $t_{gt}$ is smaller than $\tau_L$, i.e., $\left \| t_c - t_{gt} \right \| \le  \tau_L$, and considered to be false for $\left \| t_c - t_{gt} \right \| \ge  \tau_L$. Here, we use $\tau_L = 5m$ for our collected HALV dataset. To generate the PR curves of VINS-Fusion, ORB-SLAM3, and Lin et al. \cite{lin_topology_2021}, we vary a threshold on the similarity score for the DBoW2 and a threshold on the matching edit distance of \cite{lin_topology_2021}.

Fig. \ref{fig:dis_and_rot} depicts the distance and rotation error for 26 loop detection attempts conducted on our collected dataset. The horizontal axis represents the amount of ground truth rotation between the pairs of query frames and their corresponding loop frames. The results show that while VINS-Fusion can only work under the condition of camera rotation being less than 50 degrees, our method can work robustly even when the camera rotation is greater than 125 degrees.

\begin{table*}[t] \caption{Camera Pose Estimation Error on the HALV Benchmark. We evaluate absolute pose error which is presented as translational RMSE. w/o refinement: without refined object poses.}\label{table:2}
\centering
\renewcommand\arraystretch{1.2}
\setlength{\tabcolsep}{2pt}{
\begin{tabular}{l|ccc|ccc|ccc|ccc}
\hline
\multirow{2}{*}{Method} &
  \multicolumn{3}{c|}{Case 01} &
  \multicolumn{3}{c|}{Case 02} &
  \multicolumn{3}{c|}{Case 03} &
  \multicolumn{3}{c}{Case 04} \\
                & MSE(m) & STD(m) & Max(m) & MSE(m) & STD(m) & Max(m) & MSE(m) & STD(m) & Max(m) & MSE(m) & STD(m) & Max(m) \\ \hline
ORB-SLAM3\cite{ORBSLAM3_TRO}        & 1.989  & 0.494  & 2.666  & 3.604  & 1.551  & 5.558  & 3.538  & 1.465  & 5.614  & 0.626  & 0.199  & 0.998   \\
VINS-Fusion\cite{qin2019fusion}     & 1.670  & 0.825  & 3.229  & 1.336  & 0.806  & 2.702  & 0.814  & 0.418  & 1.756  & 0.720  & 0.299  & 1.130   \\
CubeSLAM\cite{cubeslam}             & 4.228  & 1.241  & 5.967  & 3.944  & 1.320  & 5.661  & 1.623  & 1.306  & 4.001  & 1.880  & 2.208  & 8.295  \\
Lin et al. \cite{lin_topology_2021} & 0.722  & 0.807  & 4.151  & 0.931  & 0.472  & 1.948  & 3.820  & 2.869  & 12.66  & 0.600  & 0.166  & 1.115  \\ \hline
Ours w/o refinement                 & 0.660  & 0.808  & 4.126  & 1.191  & 0.516  & 2.312  & 1.490  & 1.623  & 2.489  & 0.834  & 0.210  & 1.114  \\
Ours &
  \textbf{0.501} &
  \textbf{0.289} &
  \textbf{0.914} &
  \textbf{0.550} &
  \textbf{0.231} &
  \textbf{0.867} &
  \textbf{0.368} &
  \textbf{0.141} &
  \textbf{0.889} &
  \textbf{0.161} &
  \textbf{0.069} &
  \textbf{0.303} \\ \hline
\end{tabular}}
\end{table*}

\subsection{Pose Localization Performance} 
\label{exp: localization}
In this section, the proposed system w/ and w/o refinement are compared with the traditional visual, object-based SLAM systems in a few large viewpoint scenes from the HALV dataset. The reason for the system w/o refinement is to demonstrate that the accuracy of object poses is highly relevant to the performance of the loop closure detection.

We conduct experiments to localize poses and perform comparative analyses. To analyze the localization accuracy, we align the camera trajectories with ground truth using a similarity transformation \cite{horn_closed-form_1988} and calculate the absolute translation error (ATE). As shown in Fig. \ref{fig:traj}, our proposed system successfully performs loop detection and reduces cumulative camera trajectory errors compared to different visual SLAM systems. However, due to significant differences in image appearance at the ends of cases, ORB-SLAM3 and VINS-Fusion failed to detect loops and correct the drift.

The results are presented quantitatively in Table \ref{table:2}, showing the mean square error (MSE), standard deviation (STD), and max error (Max). The results demonstrate that the proposed system achieves higher accuracy in four cases with a mean MSE of 0.398m, a mean STD of 0.128m, and a mean Max of 0.743m. Additionally, we conduct ablation studies on the 'w/o refinement' version of our method to demonstrate the effectiveness of the 3D object pose refinement component. As seen from the table and figures, our proposed system significantly improves camera pose localization accuracy compared to the past feature-based and object-based methods in all cases.

\begin{table}[t]\caption{Runtime of different system components(mSec) }\label{table:3} 
\centering 
\begin{tabular}{cc}
\hline
\renewcommand\arraystretch{1.2}
Tasks                       & Runtime       \\ \hline
Data Association            & 35.2          \\
Object Optimization         & 4.3           \\
Loop Detection              & 1.8           \\
Drift Correction            & 48.4          \\ \hline
\end{tabular}
\end{table}

\subsection{Time Analysis} 
We conducted all experiments on a computer equipped with an Intel i7-9700 CPU operating at a speed of 3.00GHz and 16GB of RAM. The runtime is dependent on the number of images and the complexity of the scenes in the dataset. To determine the runtime of different system parts, we selected the image sequence in case 1. The effectiveness of the object detector is highly dependent on the computational power of the GPU and the complexity of the CNN model. As there are many lightweight methods to achieve real-time performance on a CPU, we will not discuss them here. Table \ref{table:3} shows the runtime for data association, object optimization, and drift correction, which includes pose graph optimization, global bundle adjustment, and map fusion. The tabulated data demonstrates that our system can operate in real-time, and the integration of object features has a negligible impact on the computational overhead of the system.

\section{Conclusion} \label{sec:conclusion} 
This paper proposed a novel loop closure detection method based on object-level semantic mapping. This method first extracts high-level semantic features of the environment by identifying objects, and constructs an object-level semantic map based on a novel data association method. Then, objects' neighborhood topological structures are matched based on the spatial layout and semantic properties of the graphs. Finally, it aligns the correspondence objects and jointly optimizes the poses of cameras and objects in an object-level pose graph optimization. The proposed method is evaluated in comparison with state-of-the-art methods on the KITTI dataset and the HALV dataset. The results demonstrated that the proposed method can achieve more accurate and robust camera localization performance, especially in the cases of large viewpoint changes. In future work, we plan to explore the use of more advanced deep neural networks for object detection and segmentation to improve the accuracy and efficiency of our system. 


\section*{Acknowledgments}
This work is supported by STI 2030-Major Projects (No. 2022ZD0208700) and the National Natural Science Foundation of China (No. 62276166).

\bibliographystyle{IEEEtran}
\bibliography{ref}


\begin{IEEEbiography}[{\includegraphics[width=1in,height=1.25in,clip,keepaspectratio]{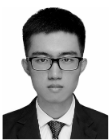}}]{Xingwu Ji}
(Graduate Student Member, IEEE) received the B.S. degree in information engineering from Xidian University, Xi'an, China, in 2019. He is currently pursuing the Ph.D. degree with Shanghai Jiao Tong University, Shanghai, China.

In 2019, he joined the Brain-Inspired Application Technology Center, Shanghai Jiao Tong University, to develop the vision brain-inspired navigation. His main areas of research interests are robotics, perception, semantic mapping.
\end{IEEEbiography}

\begin{IEEEbiography}[{\includegraphics[width=1in,height=1.25in,clip,keepaspectratio]{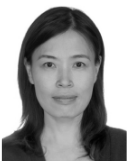}}]{Peilin Liu}
(Senior Member, IEEE) received the PhD degree from the University of Tokyo majoring in Electronic Engineering in 1998 and worked there as a research fellow in 1999. From 1999 to 2003, she worked as a senior researcher with the Central Research Institute of Fujitsu, Tokyo. 

Her research interests include low power computing architecture, application-oriented SoC design and verification, and 3D vision. She is currently a professor with Shanghai Jiao Tong University.
\end{IEEEbiography}

\begin{IEEEbiography}[{\includegraphics[width=1in,height=1.25in,clip,keepaspectratio]{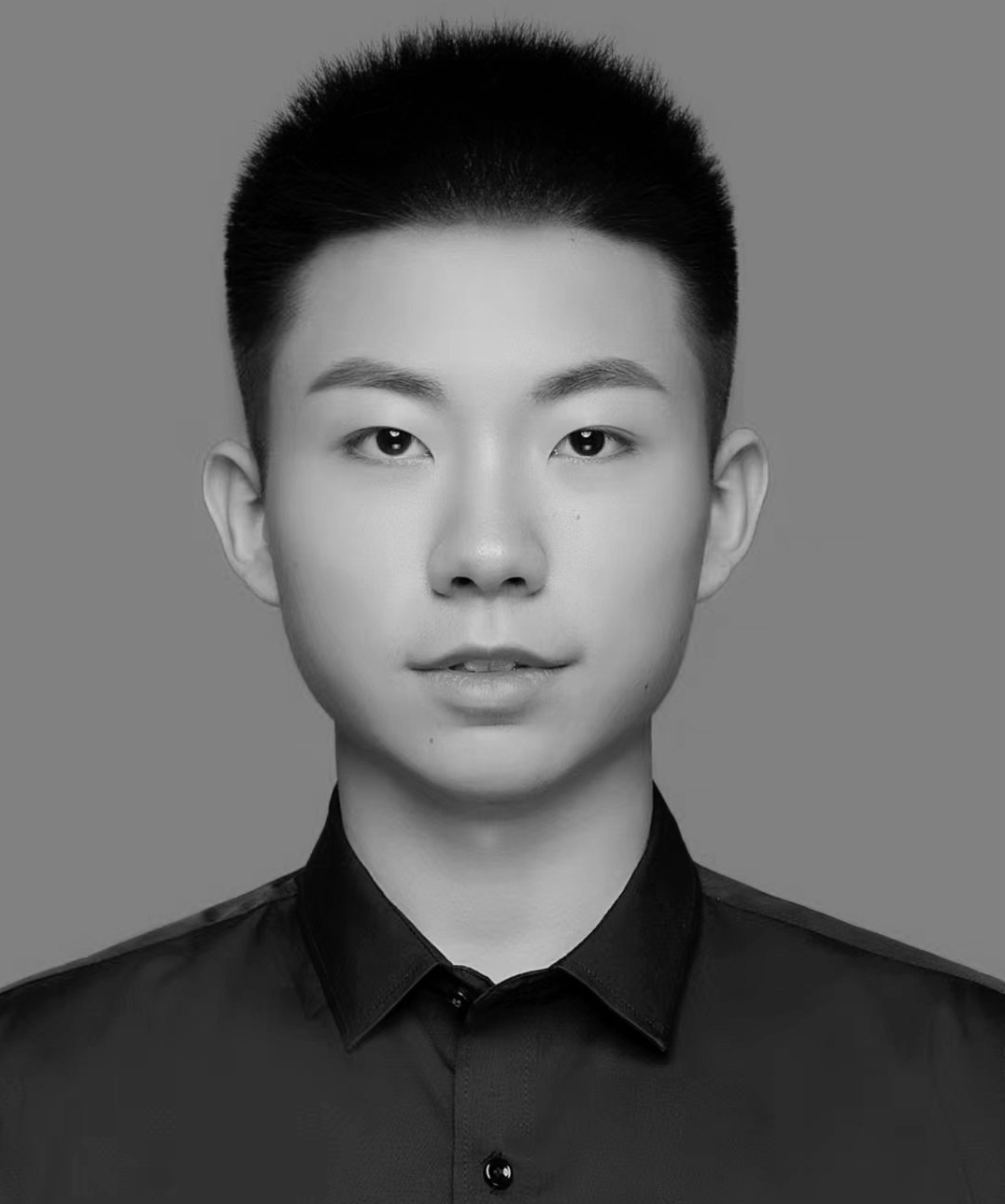}}]{Haochen Niu} received the B.S. degree from the School of Electronics and Information Engineering, Harbin Institute of Technology, Harbin, China, in 2021. He is currently pursuing the Ph.D. degree with the School of Electronic Information and Electrical Engineering, Shanghai Jiao Tong University, Shanghai, China.

His research interests include robotics, visual SLAM and computer vision.
\end{IEEEbiography}

\vfill \eject

\begin{IEEEbiography}[{\includegraphics[width=1in,height=1.25in,clip,keepaspectratio]{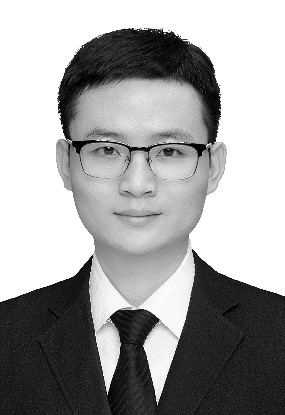}}]{Xiang Chen}
(Member, IEEE) received B.S. degree from Qingdao University, Qingdao, China, in 2013, and the master’s and Ph.D. degrees from Harbin Institute of Technology, Harbin, in 2015 and 2019, respectively, all in mechanical engineering. 

He is currently a Postdoctoral Fellow with Shanghai Jiao Tong University, Shanghai, China. His main research interests are visual servo control and path planning for robots.
\end{IEEEbiography}

\begin{IEEEbiography}[{\includegraphics[width=1in,height=1.25in,clip,keepaspectratio]{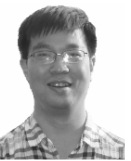}}]{Rendong Ying}
(Member, IEEE) received the B.S. degree from East China Norm University, Shanghai, China, in 1994, and the master’s and Ph.D. degrees from Shanghai Jiao Tong University, Shanghai, in 2001 and 2007, respectively, all in electronic engineering.

He is currently an Associate Professor with the Department of Electronic Engineering, Shanghai Jiao Tong University. His research area includes digital signal processing, SoC architecture, and machine thinking.
\end{IEEEbiography}

\begin{IEEEbiography}[{\includegraphics[width=1in,height=1.25in,clip,keepaspectratio]{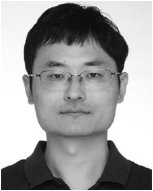}}]{Fei Wen}
(Senior Member, IEEE) received the B.S. degree from the University of Electronic Science and Technology of China (UESTC) in 2006, and the Ph.D. degree in information and communications engineering from UESTC in 2013. Now he is a research Associate Professor in the School of Electronic Information and Electrical Engineering at Shanghai Jiao Tong University. His main research interests are image processing, machine learning and robotics navigation. He serves as an Associate Editor of the International Journal of Robotics Research.

\end{IEEEbiography}

\vfill \eject

\end{document}